\documentclass{article} % For LaTeX2e
\usepackage{iclr2025_conference,times}

\usepackage{arxiv}
\usepackage{hyperref}
\usepackage{url}            % simple URL typesetting
\usepackage{booktabs}       % professional-quality tables
\usepackage{amsfonts}       % blackboard math symbols
\usepackage{nicefrac}       % compact symbols for 1/2, etc.
\usepackage{microtype}      % microtypography
\usepackage{pifont}
\usepackage{adjustbox}
\usepackage{wrapfig}
\usepackage{multirow,mathtools}
\usepackage{algorithm,algpseudocode}
\usepackage{rotating}
\usepackage{lscape}
\usepackage{longtable}
\usepackage{amssymb}
\usepackage[most]{tcolorbox}
\newtcolorbox{prompt}[2][]{colback=gray!5!white,colframe=gray!75!black,title=#2,#1}

\usepackage[font={footnotesize}]{caption}

\usepackage{multirow}
\usepackage{graphicx}
\graphicspath{{imgs/}}

\usepackage{wrapfig}

\usepackage{multirow,mathtools }

\usepackage{pifont}
\usepackage{color, colortbl}

\usepackage{blindtext}
\usepackage{lipsum}

\usepackage{multirow}
\usepackage{listings}

\usepackage{bbm}

\usepackage [english]{babel}
\usepackage[autostyle, english = american]{csquotes}

\usepackage{pifont}
\usepackage{url}
\usepackage[most]{tcolorbox}

\usepackage{lipsum}
\usepackage{soul}
\usepackage{xcolor}
\usepackage{wrapfig}
\usepackage{multirow,mathtools } 

\usepackage{adjustbox}
\MakeOuterQuote{"}

\usepackage{microtype}
\usepackage{graphicx}
\usepackage{subfigure}
\usepackage{booktabs} % for professional tables

\usepackage{hyperref}

\usepackage{amsmath}

\usepackage[capitalize,noabbrev]{cleveref}

\usepackage{nicefrac}       % compact symbols for 1/2, etc.

\usepackage{tablefootnote}

\usepackage{makecell}

\usepackage{pifont}

\usepackage{color, colortbl}
\definecolor{Gray}{gray}{0.93}
\definecolor{Orange}{rgb}{1,0.5,0}
\definecolor{DGray}{gray}{0.83}
\definecolor{LightCyan}{rgb}{0.88,1,1}

\definecolor{WarnREd}{rgb}{1,0.4,0.4}
\definecolor{WarnOrange}{rgb}{1,0.682,0.502}
\definecolor{WarnPink}{rgb}{0.9176, 0.7215, 0.7215}
\definecolor{GoodGreen}{rgb}{0.5019, 0.9215, 0.6039}

\usepackage[T1]{fontenc}

\definecolor{styleblue}{HTML}{504099}
\definecolor{mypurple}{HTML}{9391ff}

\definecolor{bluegray}{rgb}{0.4, 0.6, 0.8}
\definecolor{ceruleanblue}{rgb}{0.16, 0.32, 0.75}

\hypersetup{
colorlinks=true,
citecolor=ceruleanblue,
linkcolor=ceruleanblue,
urlcolor=black
}

\definecolor{darkgreen}{rgb}{0.0, 0.45, 0.0}
\definecolor{darkred}{rgb}{0.5, 0.0, 0.0}
\definecolor{darkblue}{rgb}{0.0, 0.0, 0.5}
\definecolor{darkyellow}{rgb}{0.65, 0.65, 0}
\usepackage{float}
\usepackage{amsmath,amsfonts,bm}

\def\eqref#1{(\ref{#1})}
\def\1{\bm{1}}

\DeclareMathAlphabet{\mathsfit}{\encodingdefault}{\sfdefault}{m}{sl}
\SetMathAlphabet{\mathsfit}{bold}{\encodingdefault}{\sfdefault}{bx}{n}

\newcommand{\E}{\mathbb{E}}

\newcommand{\KL}{D_{\mathrm{KL}}}

\DeclareMathOperator*{\minimize}{\text{minimize}}

\newcommand{\btheta}{{\boldsymbol{\theta}}}

\newcommand{\bphi}{\boldsymbol\phi}

\definecolor{SL_color}{rgb}{0.858, 0.188, 0.478}

\title{Breaking Memorization Barriers in LLM Code Fine-Tuning via Information Bottleneck for Improved Generalization}

\author{ 
Changsheng Wang$^{\dag}$\thanks{This work was initiated during an internship at Intel, and part of it was completed at OPTML after the internship.} ~~ 
Xin Chen$^{\S}$~~ 
\textbf{Sijia Liu}$^{\dag,\ddag}$ ~~ 
Ke Ding$^{\S}$ \\
$^\dag$The OPTML Lab, Michigan State University \\
$^\S$Intel\\
$^\ddag$IBM Research
}

\date{}
\begin{document}

\pagestyle{fancy}
\fancyhf{}
\cfoot{\thepage}

\maketitle

\begin{abstract}
% In this paper, we identify an overlooked phenomenon in the supervised fine-tuning (SFT) of large language models (LLMs) for code generation. We observe that when memorized data are reintroduced into the fine-tuning dataset, they block the model from effectively learning new knowledge from unseen examples, thereby limiting the expected performance gains from fine-tuning. %We term this phenomenon the \textbf{memorization barrier}
% We refer to this phenomenon as memorization barrier. Our analysis shows that the memorization barrier presents a major challenge to the effective adaptation of LLMs in domain-specific code generation. To address this issue, we propose a novel regularization approach, the \textbf{Information Bottleneck~(IB) loss}, grounded in the information bottleneck principle. In addition, we introduce an improved evaluation metric, $\mathrm{Pass@}k^{(m)}$, for fairer and more robust assessment of code generation performance. Extensive experiments demonstrate that IB-guided regularization mitigates the memorization barrier, facilitates domain knowledge learning, and achieves substantial performance improvements on two widely used code generation benchmarks.

Adapting pretrained large language models (LLMs) to code domains via supervised fine-tuning (FT) has been commonly used for code generation. However,  we identify a previously underappreciated failure mode, the \emph{memorization barrier}, where strong memorization of downstream code data in the base model could trap optimization and prevent the standard FT from effectively acquiring new, generalizable code knowledge. To overcome this barrier, we propose the \emph{information bottleneck (IB)-guided fine-tuning}, termed IB-FT, which applies an IB penalty on hidden representations of the code data to compress spurious, memorized features while preserving task-relevant information. Extensive experiments on two code benchmarks (OriGen and Evol-CodeAlpaca-V1) show that IB-FT substantially alleviates the memorization barrier, improves top-1 performance (Pass@$1$), and yields far more stable gains under the stricter multi-sample metric \text{Pass@}$k^{(m)}$ (a problem counts as solved only if at least $m$ of $k$ samples pass unit tests) compared with conventional FT.

\end{abstract}

\iffalse 
We identify an overlooked phenomenon in the supervised fine-tuning (FT) of large language models (LLMs) for code generation: the \textit{memorization barrier}. We show that many fine-tuning examples are already memorized by pretrained LLMs, which prevents models from effectively acquiring new knowledge and limits the generalization gains from FT. To address this issue, we propose an Information Bottleneck-regularized fine-tuning (IB-FT) framework that compresses spurious memorized features while preserving task-relevant information. Furthermore, we introduce a stricter evaluation metric, Pass@$k^{(m)}$, to more reliably assess generation robustness. Experiments on two diverse code generation benchmarks demonstrate that IB-FT consistently alleviates the memorization barrier, yields stronger domain adaptation, and achieves substantial improvements over standard fine-tuning. 
\fi

\section{Introduction}
\label{sec:intro}

Code generation sits at the nexus of AI and software engineering, driving rapid advances in both research and industry~\citep{nijkamp2023codegen,hou2024large,thakur2024verigen,huynh2025large,li2022competition}. By automating programming tasks, LLMs can boost developer productivity, lower engineering costs, and increase accessibility, contributions already translate to billions in annual economic value and are expected to grow as adoption widens~\citep{daniotti2025using}. 

Yet, rather than training models from scratch on specialized code corpora, practitioners typically fine-tune pretrained LLMs on downstream code datasets because this approach is computationally cheaper and more accessible~\citep{roziere2023code,beckmann2004runtime,wang2022no}. Nevertheless, adaptation remains challenging: models must acquire both programming syntax and compositional semantics while avoiding overfitting to dataset idiosyncrasies and excessive memorization of pretraining  data~\citep{mathews2024test,xu2022ide,fakhoury2024llm}. 
In this work,  we show that code fine-tuning is often brittle: fine-tuned models can exhibit a large gap between greedy decoding (Pass@$1$) and sampling-based metrics (Pass@$k$), \textit{i.e.}, a correct program may appear among multiple ($k$) sampled generations while the one-time generation by Pass@$1$ remains incorrect; We defer further details to Fig.\,\ref{fig:motivation_limitation_FT} (Sec.\,\ref{sec:preliminary_motivation}). This evidence exposes the limited effectiveness of current code fine-tuning practices and motivates our question: 
\begin{tcolorbox}[before skip=2mm, after skip=0.0cm, boxsep=0.0cm, middle=0.0cm, top=0.05cm, bottom=0.05cm, boxrule=0.6pt]
\begin{center}
\textit{\textbf{(Q)} {What causes the ineffectiveness of LLM code fine-tuning, and how can fine-tuning be modified to reliably improve generalization?}}
     \end{center}
\end{tcolorbox} 
\vspace*{2mm}

To tackle \textbf{(Q)}, we first investigate why code fine-tuning is not effective to acquire new code knowledge, examining the interaction between fine-tuning data and the pretrained model through the lens of \textit{memorization}. We identify a ``\textbf{\textit{memorization barrier}}'': the base model already strongly memorizes the fine-tuning code data, trapping optimization in a region that the standard fine-tuning objective cannot escape. 
Although prior work has identified memorization as a core challenge for LLM-based code generation~\citep{chen2025memorize,yang2024unveiling,al2024traces}, most studies treat memorization as a privacy or contamination problem, either by detecting leakage against pretraining or fine-tuning corpora~\citep{carlini2022quantifying,zeng2023exploring} or by showing that test-set contamination in pretraining inflates evaluation scores~\citep{deng2024investigating,dong2024generalization,golchin2025data,wang2025vericontaminated,riddell2024quantifying}. In contrast, we adopt a cross-setting perspective and show that strong memorization of the fine-tuning dataset already present in the base model (prior to adaptation) creates a ``memorization barrier''\footnote{\label{foot:MB_general}
Although introduced for code generation, the memorization barrier likely extends to other domains where pretrained models already memorize downstream data, thereby hindering effective adaptation.} that traps optimization and prevents effective, generalizable fine-tuning.

\begin{figure}[htb] 
%\vspace*{0mm}
\centering  
\begin{tabular}{cc}
\raisebox{3mm}{\includegraphics[width=0.28\textwidth]{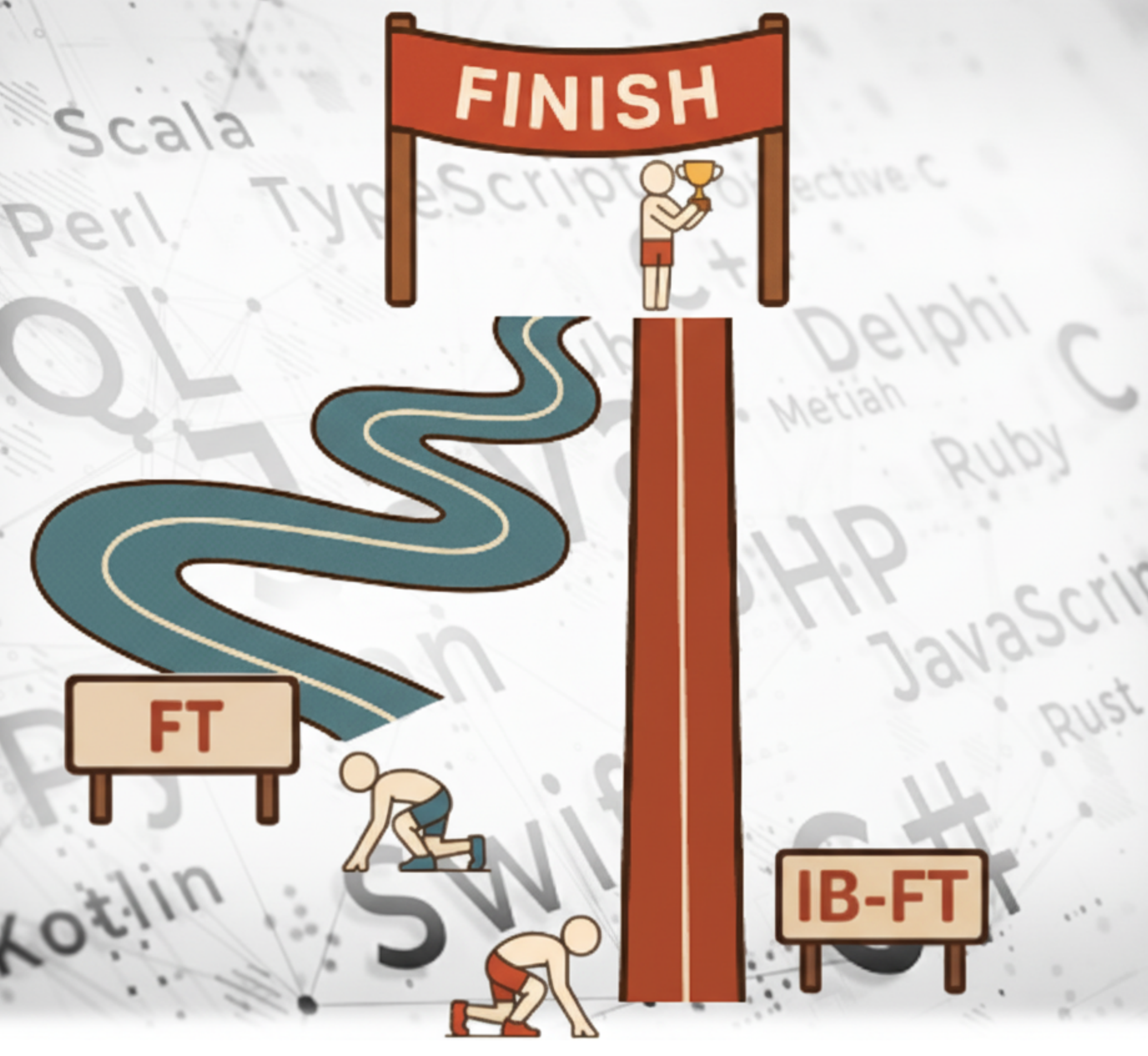}}
&
%\hspace*{-4mm}
\includegraphics[width=0.30\textwidth]{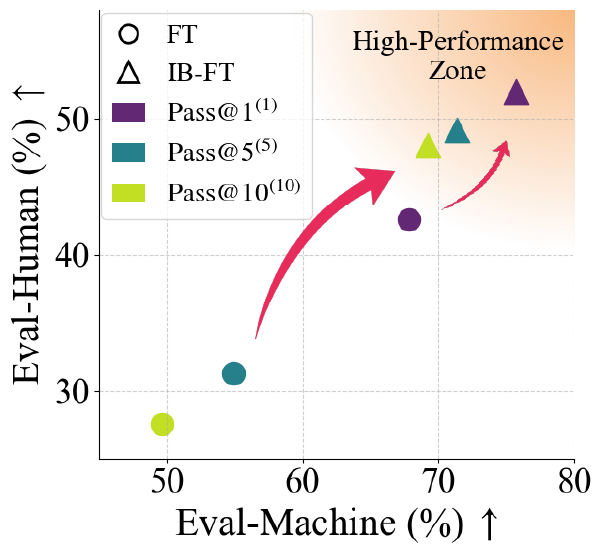}\\
  {\small (a) Code fine-tuning race }  &
  {\small (b) Fine-tuning on OriGen} 
\end{tabular}
\vspace*{0mm}
\caption{\small{
Schematic overview of IB-FT versus conventional FT and a representative performance comparison when fine-tuning on OriGen. 
(a) A code fine-tuning ``race'': conventional FT (left) appears to advance quickly by memorization but follows a winding, less-generalizable path, whereas IB-FT (right) progresses along a straighter trajectory toward the ``winning'' generalization. 
(b) Test-time performance for {DeepSeek-Coder-7B-Instruct-v1.5}, reported on Eval-Human and Eval-Machine using Pass@$k^{(m)}$ with $m=k\in\{1,5,10\}$ (a problem is counted only if all $k$ samples pass). IB-FT consistently attains higher, more stable accuracy and occupies the high-performance region compared to FT.
}}
\label{fig:motivation_intro}
\vspace*{-4mm}
\end{figure}

To overcome the memorization barrier, we also propose a novel fine-tuning (FT) strategy, termed \emph{IB-regularized fine-tuning} (IB-FT), motivated by the information bottleneck~(IB) principle~\citep{tishby2015deep,saxe2019information}. IB-FT applies an IB regularizer to hidden representations to compress prediction-irrelevant (spurious) features that arise from memorization bias. By constraining representation capacity and reshaping the data representation distribution, IB-FT reduces the dominance of heavily memorized samples and promotes more uniform learning across the code dataset, improving domain adaptation and generalization to unseen code. Intuitively, consider two runners: one starts early but follows a winding, indirect path (standard FT, which appears ahead by memorization yet is slowed by detours), while the other starts at the official line but runs straight to the finish (IB-FT, which, despite a later start, reaches the goal more directly). See \textbf{Fig.\,\ref{fig:motivation_intro}} for an illustrative schematic and performance comparison.

We summarize our main contributions below:

$\bullet$ We introduce and formalize the {memorization barrier} problem, showing via systematic analysis that strong pre-existing memorization of fine-tuning examples can critically limit the effectiveness of code fine-tuning.

$\bullet$ We recast fine-tuning through the {IB (information bottleneck)} lens and propose an IB-based regularizer (IB-FT) that compresses spurious features, alleviates the memorization barrier, and promotes more effective code fine-tuning.

$\bullet$ We present extensive empirical results validating the memorization barrier phenomenon and benchmark IB-FT on two diverse code datasets, OriGen~\citep{cui2024origen} and Evol-CodeAlpaca-V1~\citep{luo2024wizardcoder}, showing that IB-FT consistently improves generalization, including under stricter evaluations such as Pass@$1$.

\section{Related Work}

\noindent \textbf{LLM code fine-tuning.} 
Prior work on code fine-tuning has improved model performance via data-side interventions, \textit{e.g.}, synthetic or augmented corpora~\citep{luo2023wizardcoder, li2023instructcoder}, and method-side advances, \textit{e.g.}, multi-task FT, self-alignment, and parameter-efficient tuning ~\citep{ma2023training,wei2024selfcodealign,liu2024mftcoder,zhuo2024astraios}. However, these efforts largely ignore memorization in the training data, preventing the potential of code datasets from being fully exploited.

% Prior work on code fine-tuning  mainly improves performance through data-side augmentation~\citep{luo2023wizardcoder, li2023instructcoder}, such as adding comments~\citep{song2024code}, incorporating structural signals~\citep{wu2024structure}, or harmonizing multi-source corpora~\citep{song2024alchemistcoder}, and through method-side modifications~\citep{ma2023training,wei2024selfcodealign}, including multi-task SFT~\citep{liu2024mftcoder}, parameter-efficient tuning strategies~\citep{zhuo2024astraios}. 
\noindent \textbf{Information bottleneck~(IB) in LLMs.} 
The information bottleneck (IB) principle~\citep{tishby2015deep,shwartz2017opening,yang2025exploring} prescribes that latent representations should retain task-relevant information while discarding irrelevant details. Prior IB-based work has applied this idea to reasoning, calibration, security, and representation analysis, emphasizing compression or mutual-information estimation~\citep{chen2024learning,li2025calibrating,liu2024protecting,lei2025revisiting}. By contrast, to the best of our knowledge, we are the first to apply IB-guided fine-tuning in code generation, specifically to suppress spurious memorized features during adaptation.

\section{Preliminaries and Challenges in LLM Code Fine-tuning}
\label{sec:preliminary_motivation}
In this section, we introduce the preliminaries of LLM code fine-tuning, the primary task under study, and highlight its central challenge---the limited effectiveness of fine-tuning in achieving generalization on code generation tasks. We then present the core problems of our interest, serving as the focus of the subsequent sections.

% \paragraph{Preliminaries on LLM code fine-tuning.}
\noindent \textbf{Preliminaries on LLM code fine-tuning.} 
LLMs are pretrained on massive and heterogeneous corpora, which equip them with broad linguistic and reasoning capabilities. 
While such pretraining confers strong {general} abilities, it often falls short in providing the {specialized} expertise required for code-related tasks, \textit{e.g. reasoning beyond code syntax~\citep{jain2023swebench}, Python code generation~\citep{le2022coderl}, domain-specific adaptation for Verilog code generation~\cite{thakur2024verigen,zhao2024mage}}.  Instead of training an LLM from scratch on code-centric corpora, the conventional approach is to adapt a pretrained model through \emph{code fine-tuning}, where specialized code datasets are used to align the model with the target code distribution~\citep{cui2024origen, wei2025vericoder}.

% SL_fixed: add reference
 
Formally, given a pretrained model with parameters $\btheta_0$ and a target-domain code dataset 
$\mathcal{D}_{\text{code}}=\{(x,y)\}$, 
where $x$ denotes the input (\textit{e.g.}, problem description) and $y$ denotes the target output (\textit{e.g.}, solution code), 
the fine-tuning objective is to minimize the negative log-likelihood:
\begin{align}
%\hspace*{-3mm}
 \begin{array}{l}
  \displaystyle \minimize_{\btheta}   ~
  %{\ell}(\btheta) 
%\Def
\mathbb{E}_{(x,y)\sim \mathcal{D}_{\text{code}}}
\big[ -\log p_{\btheta}(y \mid x) \big],
\end{array}
%\hspace*{-3mm}
\label{eq:ft}
\end{align}
where $\btheta$ are the model parameters initialized from $\btheta_0$. 
LLM-based code models drive applications such as code completion, program synthesis, and translation. Their potential to enhance productivity, reduce errors, and democratize programming underscores the need for effective code fine-tuning.

As for Verilog code generation evaluation, \textit{VerilogEval} is commonly used benchmark~\citep{liu2023verilogeval}, which consists of two complementary subsets. \textit{Eval-Machine} refers to automatic evaluation, where generated code is executed against unit tests or reference outputs~\citep{liu2023verilogeval}. \textit{Eval-Human}, in contrast, relies on human annotators to assess the quality of generated code in terms of correctness, clarity, and style~\citep{liu2023verilogeval}. The benchmark uniformly adopts the well-known Pass@$k$ metric, which measures the probability that at least one of the top-$k$ generated samples passes the unit tests~\citep{chen2021evaluating}.

% In practice, Eval-Machine is typically employed for large-scale benchmarking due to its scalability and objectivity, while Eval-Human provides a complementary check when test suites are insufficient or when code quality requires deeper inspection.

% \SL_fixed{[You can also talk about evaluation here, e.g., pass@k, and Eval-Human and Eval-Machine (complementary? differences?)] E.g., \textit{Eval-Machine} refers to automatic evaluation, where generated code is executed against unit tests or reference outputs [refs]. \textit{Eval-Human} relies on human annotators to judge the quality of generated code in terms of correctness, clarity, and style [refs].  In practice, Eval-Machine is typically used for large-scale benchmarking, while Eval-Human serves as a complementary check when test suites are insufficient or code quality needs deeper inspection.}

\begin{figure}[htb] 
%\vspace*{0mm}
\centering  
\begin{tabular}{cc}
\includegraphics[width=0.3\textwidth]{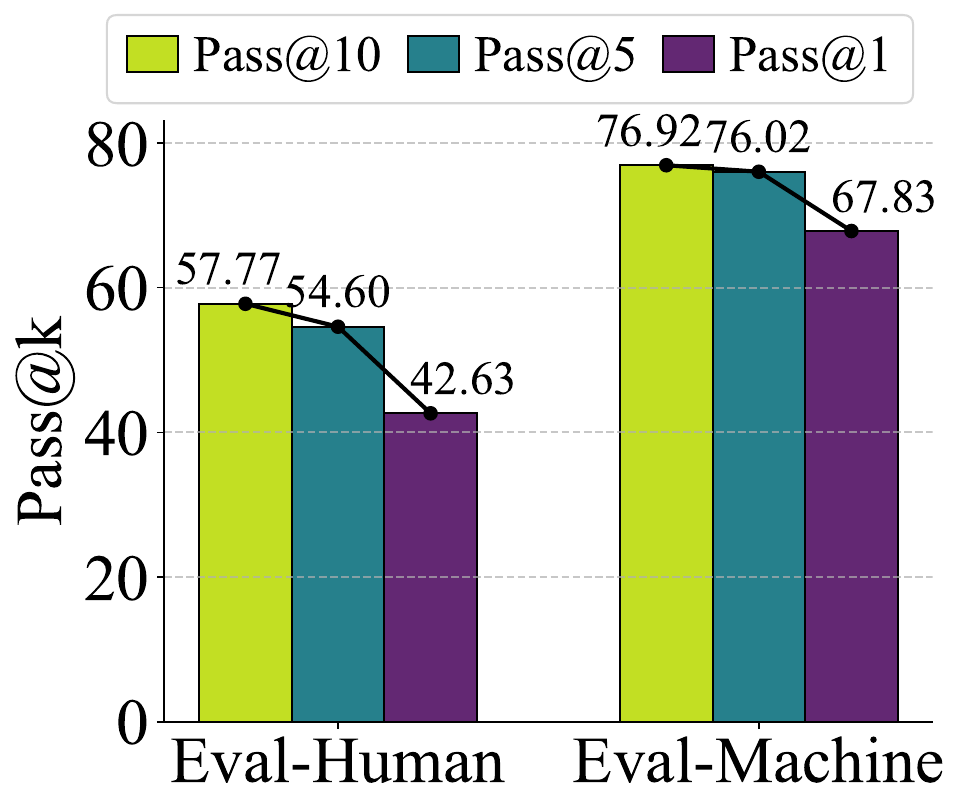} &
\includegraphics[width=0.31\textwidth]{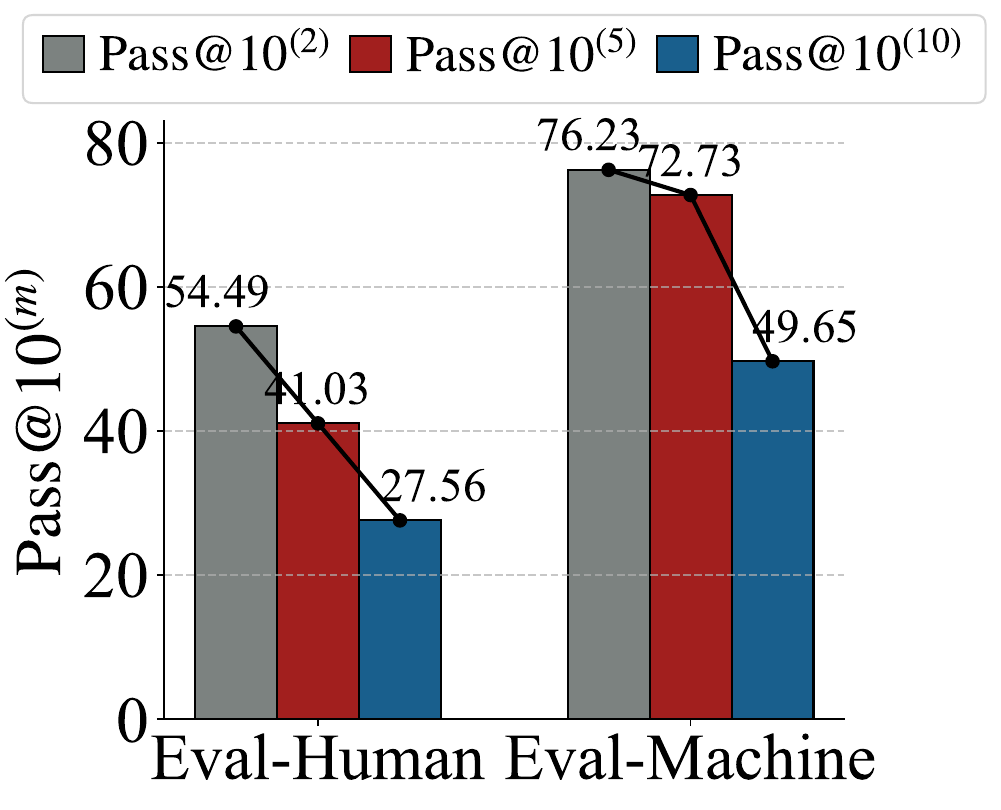}\\
  % {\footnotesize (a) }  &
  % {\footnotesize (b) } 
\end{tabular}
\vspace*{-3mm}
\caption{
\small{Performance of DeepSeek-Coder-7B-Instruct-v1.5 fine-tuned on the OriGen dataset, evaluated on the Eval-Human and Eval-Machine. (a) Pass@$k$ results for $k \in \{10, 5, 1\}$, measuring the probability that at least one of the top-$k$ generations passes the test cases. (b) Pass@$k^{(m)}$ results with $k=10$ and $m \in \{2,5,10\}$, requiring multiple successful generations among the ten samples to reflect robustness.}
}
\label{fig:motivation_limitation_FT}
\vspace*{-3mm}
\end{figure}

% \paragraph{An illusion of effectiveness in code fine-tuning.}

\noindent \textbf{An illusion of effectiveness in code fine-tuning.}
Before our study, it was commonly believed that code fine-tuning delivers acceptable performance, particularly under the \textit{Pass@$k$} metric, which measures the probability that \textit{at least one} of the top-$k$ generated outputs from an LLM is correct. 
However, we find that this might be an illusion of effectiveness: (1) under greedy decoding (\textit{i.e.}, $k=1$), the fine-tuned LLM can experience a sharp accuracy drop, and (2) even for larger $k$, most generations actually fail despite one occasionally passing.

\textbf{Fig.\,\ref{fig:motivation_limitation_FT}} validates the above two limitations by fine-tuning the pretrained DeepSeek-Coder-7B-Instruct-v1.5 on the OriGen, with test-time generalization performance evaluated on Eval-Human and  Eval-Machine. 
As shown in Fig.\,\ref{fig:motivation_limitation_FT}(a), the fine-tuned model performance, evaluated by both \textit{Eval-Machine} and \textit{Eval-Human}, exhibits a significant accuracy drop at Pass@$1$. 
This indicates that the fine-tuned model lacks certainty under greedy decoding, unless one leverages the strength of probabilistic generation. 
Furthermore, to probe beyond the standard Pass@$k$, we introduce a complementary metric \textit{Pass@$k^{(m)}$}, which measures the probability that \textit{at least $m$} of the top-$k$ generated outputs are correct. 
Note that Pass@$k^{(1)}$ reduces to the classic Pass@$k$ evaluation. 
As shown in Fig.\,\ref{fig:motivation_limitation_FT}(b), the accuracy under Pass@$k^{(m)}$ decreases sharply as $m$ increases, revealing that even when allowing multiple generations (\textit{e.g.}, $k=10$), most outputs are still incorrect; for instance, $m=5$ requires half of the generations to succeed, yet performance drops drastically in this setting.

% \SL_fixed{[be consistent if you used Pass@$k^{(n)}$ in all texts, image captions, and figures]}

% \paragraph{Problem statement.}
\noindent \textbf{Problem statement.}
As shown in Fig.\,\ref{fig:motivation_limitation_FT}, achieving effective code fine-tuning for LLMs is far from trivial, with root causes largely overlooked in prior work. 
To address this gap, we focus on two key questions in this study: 
\textit{\textbf{(Q1)} What causes the difficulty of effective code fine-tuning for LLMs?} 
\textit{\textbf{(Q2)} How can the existing fine-tuning protocol in \eqref{eq:ft} be advanced to achieve improved effectiveness?}

In the rest of the paper, we address (Q1) by identifying the ``memorization barrier'' in LLM code fine-tuning, where excessive memorization of fine-tuning data by the pretrained LLM can \textit{hinder} effective adaptation (see Sec.\,\ref{sec:Method1}). 
We then address (Q2) by tackling this barrier through an information bottleneck perspective, enforcing \textit{equality} across data with different memorization levels so that the fine-tuner learns from the most informative aspects of the data (see Sec.\,\ref{sec:IB_method}).

\section{The ``Memorization Barrier'' in LLM Code Fine-tuning}
\label{sec:Method1}
In this section, we identify and formalize the ``\textit{memorization barrier}'', introduced here for the first time, as a potential root cause of the challenges in code fine-tuning.

% \paragraph{The memorization lens on (in)effective code fine-tuning.}
\noindent \textbf{The memorization lens on (in)effective code fine-tuning.}
As illustrated in Fig.\,\ref{fig:motivation_limitation_FT}, the  code fine-tuned LLM does not appear to gain much from the code dataset, as evidenced by its sharp accuracy drop in Pass@$1$ and in Pass@$k^{(m)}$ for large $m$. 
This raises the question of whether the limitation stems from the quality of the data themselves or from the fine-tuning process failing to fully exploit the code data. 
In this work, we assume no control over data curation and instead focus on improving the fine-tuning process. 

Our hypothesis is that effective code fine-tuning should enable the model to acquire new domain knowledge beyond pretraining, which motivates a deeper examination of the relationship between fine-tuning data and the pretrained LLM.
We approach this from the perspective of \emph{memorization} \citep{shi2023detecting,biderman2023emergent,leybzon2024learning, kiyomaru2024comprehensive}: Given that the pretraining corpus is massive and may partially overlap with the target domain, we examine whether a non-trivial portion of the fine-tuning data has already been memorized, and how this entanglement, ignored in the conventional fine-tuning protocol \ref{eq:ft}, underlie the observed limitations.

% \begin{figure}[htb] 
% \vspace*{0mm}
% \centering  
% \begin{tabular}{cc}
% \hspace*{-6mm}
% \includegraphics[width=0.25\textwidth]{figure/min20_OriGen_v3.pdf} &
% \hspace*{-4mm}\includegraphics[width=0.25\textwidth]{figure/min20_CodeAlpaca_v3.pdf}\\
%   {\scriptsize (a) OriGen}  &
%   {\scriptsize (b) Evol-CodeAlpaca-V1} 
% \end{tabular}
% \vspace*{-3mm}
% \caption{\small{Downstream data already memorized during pretraining. Memorization patterns measured by \texttt{Min-K\% Prob} ($K=20$) using DeepSeek-Coder-7b-Instruct-v1.5 and Llama-3 8B as base models on two code datasets, OriGen and Evol-CodeAlpaca-V1. The x-axis shows \texttt{Min-20\% Prob} scores and the y-axis their density. For each dataset, we report three \texttt{Min-20\% Prob} distributions: $p_{0}(\mathcal{D}_\mathrm{code}; \btheta_0)$ (code dataset on the base model), $p_{1}(\mathcal{D}_\mathrm{code}; \btheta_\mathrm{code})$ (code dataset on the fine-tuned model), and $p_{2}(\mathcal{D}_\mathrm{TOFU}; \btheta_0)$ (TOFU dataset on the base model), where $p_2$ gives a weakest-memorization reference.
% }
% }
% \label{fig:mink_dis}
% \vspace*{-4mm}
% \end{figure}

\begin{figure}[htb] 
\vspace*{-1mm}
\centering  
% ==========
\includegraphics[width=0.7\textwidth]{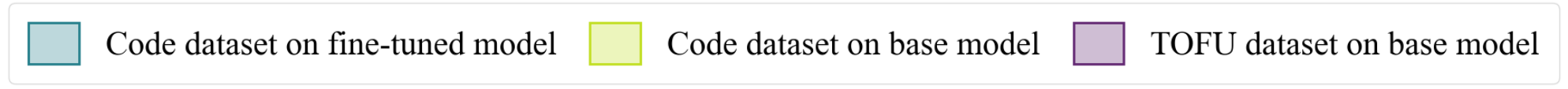} 
\vspace{0mm}

% ==========
\begin{tabular}{cc}
%\hspace*{-6mm}
\includegraphics[width=0.35\textwidth]{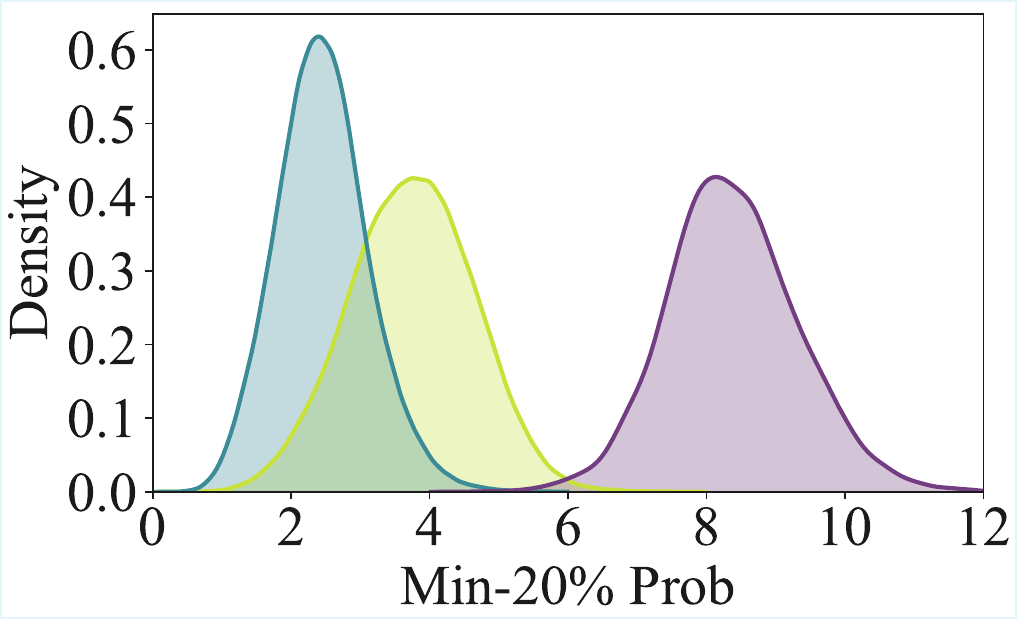} &
%\hspace*{-4mm}
\includegraphics[width=0.35\textwidth]{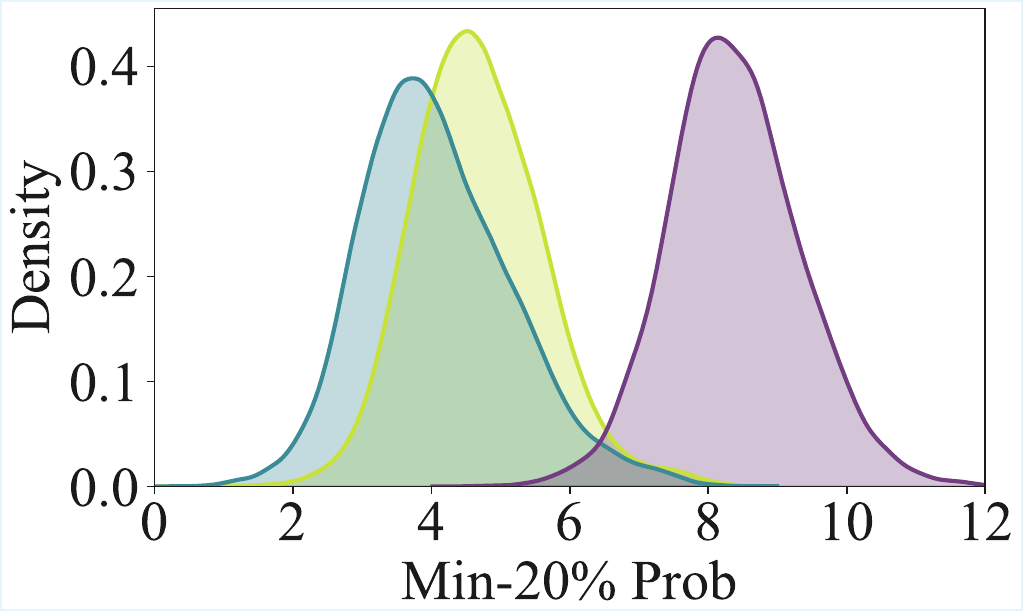}\\
  {\small (a) OriGen}  &
  {\small (b) Evol-CodeAlpaca-V1} 
\end{tabular}

\vspace*{-0mm}
\caption{\small{Downstream data already memorized during pretraining. Memorization patterns measured by \texttt{Min-K\% Prob} ($K=20$) using DeepSeek-Coder-7b-Instruct-v1.5 and Llama-3 8B as base models on two code datasets, OriGen and Evol-CodeAlpaca-V1. The x-axis shows \texttt{Min-20\% Prob} scores and the y-axis their density. For each dataset, three distributions are reported: $p_{0}(\mathcal{D}_\mathrm{code}; \btheta_0)$ (code dataset on the base model), $p_{1}(\mathcal{D}_\mathrm{code}; \btheta_\mathrm{code})$ (code dataset on the fine-tuned model), and $p_{2}(\mathcal{D}_\mathrm{TOFU}; \btheta_0)$ (TOFU dataset on the base model), where $p_2$ provides the weakest-memorization reference.}}
\label{fig:mink_dis}
\vspace*{-3mm}
\end{figure}

%The distribution of the pretrained model on each downstream dataset is compared against two references: a fine-tuned reference obtained by evaluating the same dataset with a model trained on it (labeled “Finetuned model on OriGen/CodeAlpaca”), which reflects the upper bound of memorization, and an unseen reference obtained from the synthetic TOFU dataset~\citep{maini2024tofu} (labeled “Base model on synthetic data”), which is guaranteed to be absent from the pretraining corpus and thus serves as a proxy for non-memorized content. The substantial overlap between the pretrained and fine-tuned distributions, together with their clear separation from the TOFU distribution, indicates that a significant fraction of downstream examples were already memorized during pretraining. 

Through the lens of memorization, \textbf{Fig.\,\ref{fig:mink_dis}} shows the memorization scores of the fine-tuning code datasets OriGen~\citep{cui2024origen} and Evol-CodeAlpaca-V1~\citep{luo2023wizardcoder} with respect to the base model ($\btheta_0$).
Here, memorization is assessed using the \texttt{Min-K\% Prob} method~\citep{shi2023detecting}, which measures how likely a sequence has been memorized by a model. It selects the $K\%$ tokens with the lowest predicted probabilities and computes their average negative log-likelihood (\textit{i.e.}, the prediction loss over the least likely tokens).
A lower \texttt{Min-K\% Prob} score indicates stronger memorization, as even the least likely tokens are predicted confidently, whereas a higher score suggests weaker memorization.
% (\textit{i.e.}, the negative log-likelihood)

In Fig.\,\ref{fig:mink_dis}, we present the distribution of \texttt{Min-K\% Prob} (with $K=20$)
scores for the fine-tuning code dataset (OriGen or Evol-CodeAlpaca-V1, denoted $\mathcal{D}_\mathrm{code}$) on the pretrained model ($\btheta_0$) and its fine-tuned counterpart ($\btheta_{\mathrm{code}}$), denoted as $p_{0}(\mathcal{D}_\mathrm{code}; \btheta_0)$ and $p_{1}(\mathcal{D}_\mathrm{code}; \btheta_\mathrm{code})$, respectively. For comparison, we also include $p_{2}(\mathcal{D}_\mathrm{TOFU}; \btheta_0)$, the distribution obtained from the base model evaluated on the fictitious TOFU dataset~\cite{maini2024tofu}, which is excluded from pretraining as it contains synthetic author profiles.
The rationale is that $p_{2}(\mathcal{D}_\mathrm{TOFU}; \btheta_0)$ serves as \textit{a reference distribution for the weakest memorization} of the base model, since $\mathcal{D}_\mathrm{TOFU}$ contains entirely fictitious information.
As shown in Fig.\,\ref{fig:mink_dis}, $p_{0}$ lies to the right of $p_{1}$, reflecting the stronger memorization of $\mathcal{D}_\mathrm{code}$ by the fine-tuned model $\btheta_\mathrm{code}$. This is expected, since $\btheta_\mathrm{code}$ is directly trained on $\mathcal{D}_\mathrm{code}$. More interestingly, \textit{however}, $p_{0}$ has shown substantial overlap with $p_{1}$, suggesting that \textit{the base model has already memorized much of the dataset prior to fine-tuning}. This strong pre-existing memorization becomes especially evident when contrasted with the weak memorization reference $p_{2}$.

% \SL_fixed{[The following paragraph can make it conciser as this has been mentioned in introduction by me.]}

The above reveals a \textit{new memorization phenomenon} identified from LLM code fine-tuning: {fine-tuning data points} may already be strongly {memorized by the pretrained base model} even before fine-tuning. 
We remark that this finding differs from prior literature, which has typically examined memorization from a privacy perspective (to detect leakage), either by evaluating an LLM against its pretraining data \citep{carlini2022quantifying} or by assessing a fine-tuned LLM against its fine-tuning data \citep{zeng2023exploring}. In contrast, ours introduces a \textit{cross-setting} perspective: evaluating the memorization of \textit{fine-tuning datasets} on the \textit{base model}.

% \paragraph{``Memorization barrier'' in code fine-tuning.}
\noindent \textbf{``Memorization barrier'' in code fine-tuning.}
Given the strong memorization of code fine-tuning data by the pretrained base model, we posit that this memorization underlies the limited effectiveness of LLM code fine-tuning. 
This challenge is analogous to a known challenge in nonconvex optimization, ``\textit{escaping bad local optima}'' \citep{ge2015escaping,criscitiello2019efficiently}.
Our rationale is that, due to the memorization, the base model becomes trapped in a {bad} ``local optimum'' characterized by high memorization. Consequently, the conventional fine-tuning objective \eqref{eq:ft} struggles to escape this state and converge to a better solution, \textit{i.e.}, a code fine-tuned model with improved generalization.
We refer to this issue as  ``\textbf{memorization barrier}''  and state it below. 
  
\begin{center}
\vspace*{-0mm}
	\setlength\fboxrule{0.5pt}
	\noindent\fcolorbox{black}[rgb]{0.99,0.99,0.99}{\begin{minipage}{0.96\columnwidth}
    %\vspace*{-4mm}
\textbf{Memorization barrier:} The phenomenon where LLM code fine-tuning starts from a base model $\btheta_0$ that already strongly memorizes the fine-tuning set $\mathcal{D}_{\mathrm{code}}$, placing optimization in a state the conventional fine-tuning objective struggles to escape, thereby leading to poor generalization on downstream code tasks.
	\end{minipage}}
	%\vspace*{-2.5mm}
\end{center}

\begin{figure}[!htb] 
\vspace*{0mm}
\centering  
\begin{tabular}{cc}
%\hspace*{-6mm}
\includegraphics[width=0.33\textwidth]{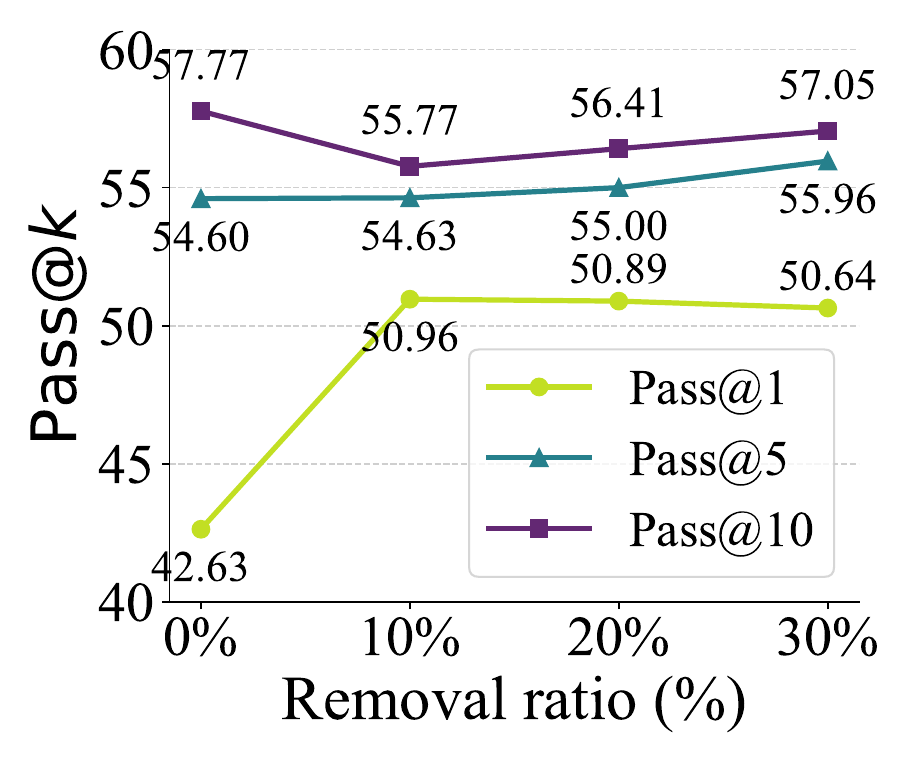} &
%\hspace*{-4mm}
\includegraphics[width=0.33\textwidth]{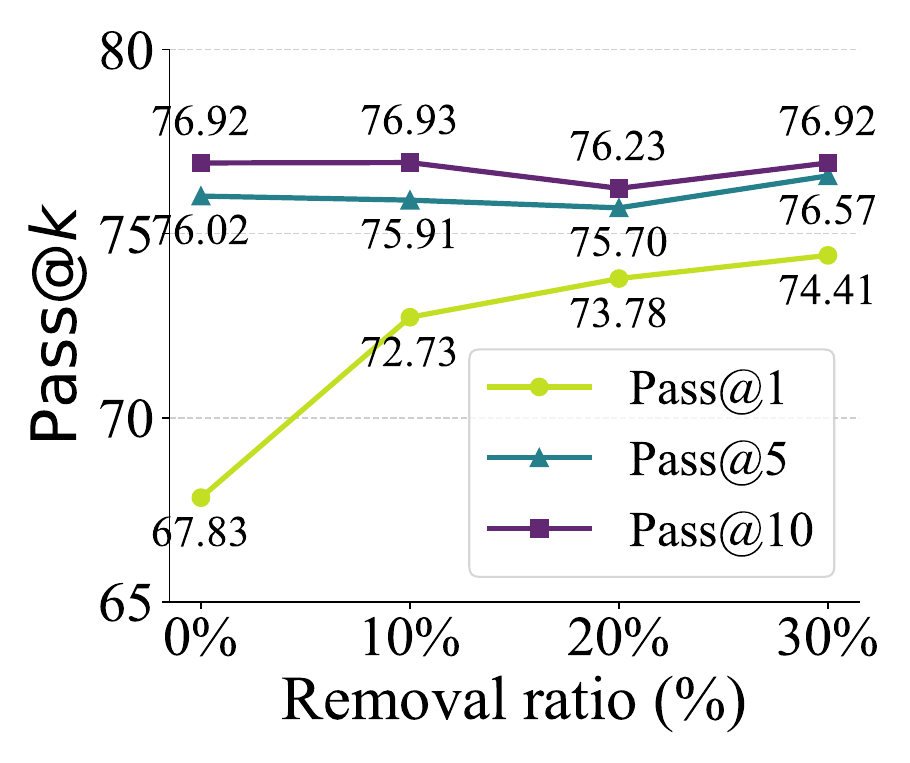} \vspace*{-2mm}\\
  {\small (a) Eval-Human}  &
  {\small (b) Eval-Machine} 
\end{tabular}
\vspace*{-1mm}
\caption{\small{Pass@$k$ performance of the fine-tuned model on the OriGen dataset using DeepSeek-Coder-7b-Instruct-v1.5, evaluated by (a) \textit{Eval-Human} and (b) \textit{Eval-Machine}~\citep{liu2023verilogeval}, with different pruning ratios of highly memorized fine-tuning data identified in Fig.\,\ref{fig:mink_dis}. The x-axis shows the removal ratio of the most memorized examples (0\% corresponds to full dataset), and the y-axis reports Pass@$k$ for $k \in \{1,5,10\}$.
}
}
\label{fig:remove_on_sft_main}
\vspace*{-4mm}
\end{figure}

\textbf{Fig.\,\ref{fig:remove_on_sft_main}} validates the memorization barrier by comparing \textit{Pass@$k$} performance of the fine-tuned model using the conventional approach \eqref{eq:ft} with different data pruning ratios, where the most memorized data points (identified in Fig.\,\ref{fig:mink_dis}) are removed from the fine-tuning set. We examine whether excluding highly memorized fine-tuning data, \textit{i.e.}, alleviating the memorization barrier at the data source, enables the conventional fine-tuning approach to yield a better code model.
The results reveal a striking pattern: removing as little as the most memorized 10\% of data yields a substantial improvement in Pass@$1$ on Eval-Human and Eval-Machine. Further, using fewer fine-tuning data points does not degrade generalization under Pass@$k$ for large $k$. These suggest that memorized data act as a \emph{memorization barrier}, capping the effectiveness of fine-tuning from the base model.

The memorization barrier reveals a new effect of memorization on the degraded generalization of LLM fine-tuning. This is different from prior studies, which have primarily examined its negative impact in the context of data contamination~\citep{deng2024investigating,dong2024generalization,golchin2025data,wang2025vericontaminated,riddell2024quantifying}, where test samples appear in the pretraining corpus, creating a false impression of ``superior'' generalization.
To tackle this memorization barrier problem, dataset pruning based on memorization appears to be a straightforward way to improve code fine-tuning as shown in Fig.\,\ref{fig:remove_on_sft_main}. However, this approach is \textit{inefficient}: (1) it requires data-wise attribution to estimate memorization, which is difficult to scale since attribution itself is computationally intensive; and (2) determining the optimal pruning ratio for improved generalization is nontrivial, as exploring all possible choices requires multiple fine-tuning runs, making the process impractical.  In the next section, we address the memorization barrier using the \textit{information bottleneck} principle, which eliminates the need for additional data attribution and pruning.

\section{Information Bottleneck–Guided LLM Code Fine-tuning}
\label{sec:IB_method}
In this section, we address the memorization barrier in LLM code fine-tuning by leveraging the information bottleneck (\textbf{IB}) as a principled form of regularization. The IB framework reduces redundant memorization, preserves task-relevant information, and reshapes representations to foster improved generalization.

% \paragraph{IB regularization to overcome the memorization barrier.} 
\noindent \textbf{IB regularization to overcome the memorization barrier.}
To address the memorization barrier identified in Sec.\,\ref{sec:Method1}, we adopt the IB principle \citep{tishby2000information,lei2025revisiting,liu2024protecting,chen2024learning}  as a remedy that requires no additional data attribution or pruning, enabling an automatic fine-tuning protocol without assuming prior knowledge of the barrier. 
Motivated by the analogy between the memorization barrier and the problem of \textit{escaping bad local optima}~\citep{ge2015escaping,criscitiello2019efficiently} discussed in Sec.\,\ref{sec:Method1}, we note that the latter can often be mitigated by \textit{random re-initialization} in nonconvex optimization~\citep{lee2016gradient}. Random initialization can be understood as restarting the optimization process, providing an equal opportunity to create a new optimization path and escape bad local optima.

Therefore, similar to random re-initialization, the rationale behind using IB is to transform differently memorized data into equally treated ones by compressing representations to discard spurious information while preserving task-relevant signals for prediction \citep{tishby2000information}.
More concretely, the IB principle provides a framework for learning representations that are both predictive and compressed. Given input $X$, target $Y$, and representation $Z$, IB seeks $Z$ that preserves task-relevant information about $Y$ while discarding redundant details from $X$. This is formulated as
\begin{align}
\begin{array}{l}
\displaystyle \minimize_{p(z|x)} \; I(X;Z) - \beta I(Z;Y),
\label{eq:ib-objective}
\end{array}
\end{align}
where $p(z|x)$ gives the conditional distribution of the latent representation $Z$ given the input $X$, and 
$I(\cdot;\cdot)$ denotes mutual information and $\beta>0$ balances prediction against compression.

% Rather than discarding data, IB compresses irrelevant details while retaining task-relevant signals for prediction:
% \begin{align}
% \begin{array}{l}
% \min \; I(X;Z) - \beta I(Z;Y),
% \label{eq:ib-objective}
% \end{array}
% \end{align}
% where $I(\cdot \,;\cdot)$ denotes mutual information, $X$ is the input, $Z$ the bottleneck representation, $Y$ is the target, and $\beta$ controls the trade-off.

%Xin change First, for compression term $I(X;Z)$,this term 
The compression term $I(X;Z)$ penalizes how much bottleneck representation $Z$ memorizes input-specific details. In LLMs, direct computation is intractable, so we follow the variational IB framework~\citep{alemi2017deep}:
\begin{align}
\begin{array}{l}
I(X;Z) \;\;\leq\;\; \E_{x \sim \mathcal{D}_\mathrm{code}}\left [ \KL  \big(q_{\boldsymbol{\phi}}(z|h_{\btheta}(x))\,\Vert\,p(z)\big) \right ],
\label{eq:compression-app}
\end{array}
\end{align}
where $\mathcal{D}_\mathrm{code}$ denotes the dataset, 
$\KL$ is the Kullback–Leibler divergence, 
$p(z)$ is a simple prior (\textit{e.g.}, $\mathcal{N}(0,I)$), 
$h_{\btheta}(x)$ denotes the hidden representation of input $x$ extracted from a designated intermediate LLM layer (\textit{e.g.}, layer 20), 
and $q_{\bphi}$ is the variational encoder (with the learnable parameters $\bphi$) that produces the representation $Z$ given $h_{\btheta}(x)$. 
The design of $q_{\bphi}$ follows the standard variational IB setting~\citep{alemi2017deep}.

Minimizing this divergence discourages $Z$ from encoding exact input patterns, thereby suppressing memorization of rare or spurious sequences. %For more details We define   xin deleted 
Hence, from \eqref{eq:compression-app}, the compression loss is given by 

\begin{align}
\begin{array}{l}
 \ell_{\mathrm{IB}}^{\mathrm{compress}}  (\btheta,\bphi) 
=   \E_{x \sim \mathcal{D}_\mathrm{code}}\left [ \KL  \big(q_{\bphi}(z  \mid h_{\btheta}(x))\,\Vert\,p(z)\big) \right ],
\label{eq:compression-final}
\end{array}
\end{align}

In addition, the prediction term $I(Z;Y)$ in \eqref{eq:ib-objective} prevents over-compression by ensuring that the bottleneck preserves information relevant for predicting $Y$. This term can be expressed by the log-likelihood of predicting $Y$ from $Z$, 

\begin{align}
\begin{array}{l}
\ell_{\mathrm{IB}}^{\mathrm{predict}}(\btheta,\bphi)
= \,\E_{(x,y)\sim\mathcal{D}_{\mathrm{code}}}
\left [ \log p_\btheta(y \mid z) \right ],
\label{eq:prediction-final}
\end{array}
\end{align}

where $z$ is sampled from $q_{\bphi}(z|h_{\btheta}(x))$.

Integrating \eqref{eq:compression-final} with \eqref{eq:prediction-final} based on \eqref{eq:ib-objective}, the proposed IB regularization loss is defined as
\begin{align}
\begin{array}{l}
\ell_{\mathrm{IB}} (\btheta,\bphi)
= \ell_{\mathrm{IB}}^{\mathrm{compress}} (\btheta,\bphi)
- \beta \ell_{\mathrm{IB}}^{\mathrm{predict}} (\btheta,\bphi),
\label{eq:IB-loss}
\end{array}
\end{align}
where $\beta$ is an empirically chosen hyperparameter that balances the tradeoff between compression and prediction, consistent with the standard IB formulation.
 % \SL_fixed{[Please remember to talk about this choice range in experiment setups.]}
%
%\SL{[how to choose $\beta$ in practice?]} \xin{The hyperparameter $\beta$ is set empirically within a reasonable range to balance compression and prediction}
% $\ell_{\mathrm{IB}}^{\mathrm{KL}}$ is the KL divergence penalty that discourages $Z$ from memorizing input-specific details, and $\ell_{\mathrm{IB}}^{\mathrm{CE}}$ is the predictive cross-entropy loss along the bottleneck path. The hyperparameter $\beta>0$ balances prediction relevance against compression within the IB term.
Furthermore, we integrate \eqref{eq:IB-loss} with the standard fine-tuning objective in \eqref{eq:ft}, yielding the \textbf{IB-regularized fine-tuning (IB-FT)}:
\begin{equation}
\begin{array}{l}
   \displaystyle \minimize_{\btheta,\bphi}  ~~ \ell_{\mathrm{FT}}(\btheta) + \alpha \, \ell_{\mathrm{IB}}(\btheta,\bphi),
\end{array}
\tag{IB-FT}
\label{eq:final-loss}
\end{equation}
where $\ell_{\mathrm{FT}}$ is the standard cross-entropy loss from \eqref{eq:ft}, and $\alpha > 0$ controls the relative strength of the IB regularization.

\noindent \textbf{Validating the effectiveness of IB through representation analysis.} Recall that the motivation for using IB lies in its ability to compress representations while retaining predictive information, thereby reducing the influence of prediction-irrelevant (\textit{i.e.}, spurious) features. This allows differently memorized data to be treated more uniformly during learning, mitigating the memorization barrier that arises from prior differences in memorization strength among code data. 

\begin{figure}[htb] 
%xin{fix the figure position avoid the late the figure float}
\vspace*{0mm}
\centering  
\begin{tabular}{cc}
%\hspace*{-6mm}
\includegraphics[width=0.3\textwidth]{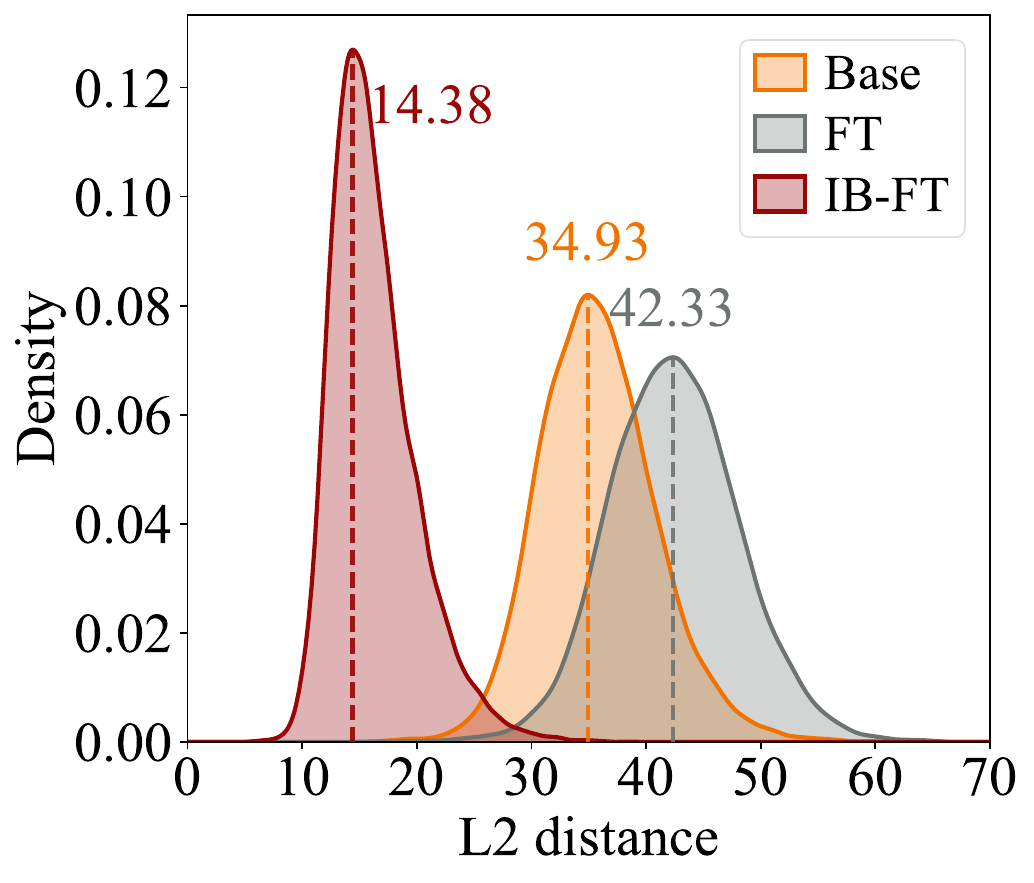} &
%\hspace*{-4mm}
\includegraphics[width=0.3\textwidth]{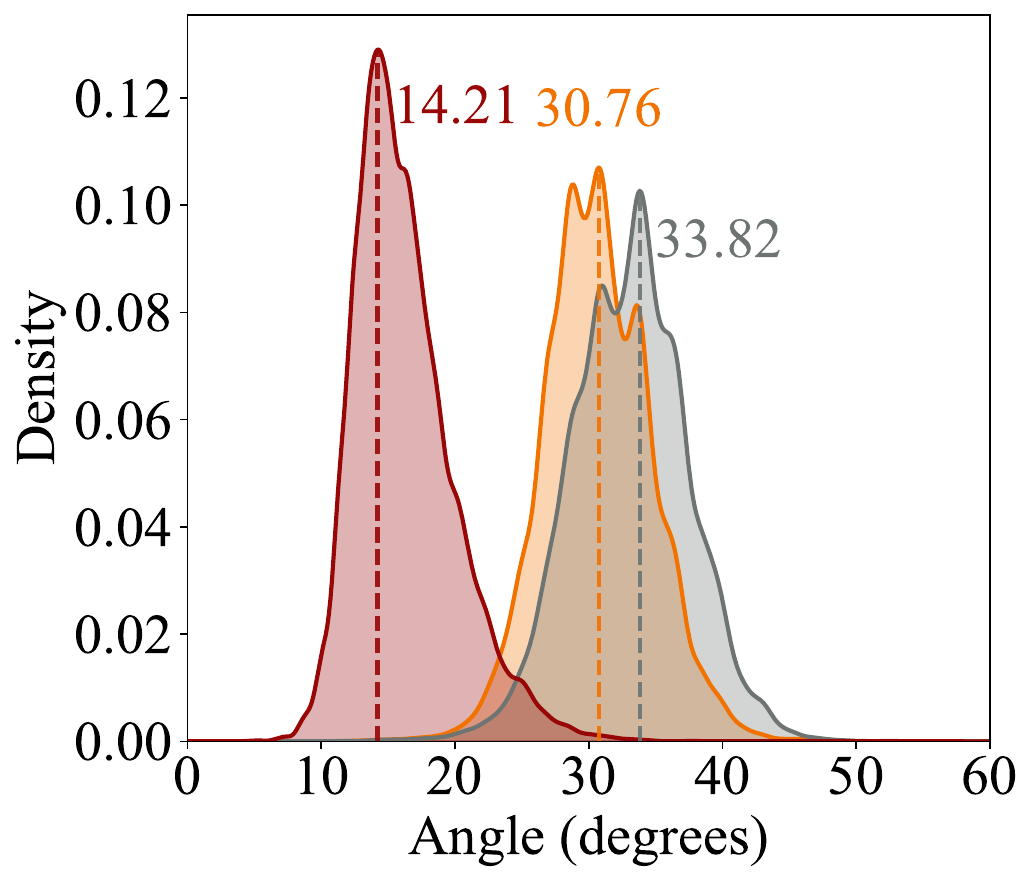} \vspace*{-2mm}\\
\end{tabular}
\vspace*{-0mm}
\caption{\small{Representation distance and angle analysis with fine-tuning data on OriGen, using the base DeepSeek-Coder-7B-Instruct-v1.5 model and its fine-tuned variants (FT and IB-FT). 
Hidden representations $h_\btheta(x)$ are extracted from the 20th layer, and paired samples are drawn between the most-memorized and least-memorized groups (identified by \texttt{Min-20\% Prob} scores in Fig.\,\ref{fig:mink_dis}) to compute (\textbf{left}) their $\ell_2$ distance distributions and (\textbf{right}) angular disparity distributions. 
Dashed vertical lines mark distinct separation patterns under different fine-tuning regimes.
}
}
\label{fig:rep_distance}
\vspace*{-4mm}
\end{figure}

To validate the effectiveness of IB-FT, we analyze the representation geometry between memorized and non-memorized examples. Using \texttt{Min-K\% Prob} scores, the dataset is partitioned into two subsets (most-memorized group and least-memorized group), and paired samples are drawn to measure $\ell_2$ distances and angular disparities between their hidden representations at layer 20. As shown in \textbf{Fig.\,\ref{fig:rep_distance}}, standard FT exaggerates the separation between the two groups relative to the base model, creating a barrier that forces the model to treat them in markedly different ways. In contrast, IB-FT substantially compresses this gap, reducing both distance and angular disparity by more than half. These results demonstrate that IB regularization yields more coherent, task-relevant representations across all data groups, thereby enabling fine-tuning to generalize more effectively.

\section{Experiments}

\subsection{Experiment Setup}
% \begin{table*}[!htb]

\noindent \textbf{Datasets and models.}
% \paragraph{Datasets and models.} 
Our experiments focus on two established datasets for LLM code fine-tuning: \textbf{OriGen}~\citep{cui2024origen} for Verilog code generation, and \textbf{Evol-CodeAlpaca-V1}~\citep{luo2024wizardcoder} for multi-language code generation. 
\underline{OriGen}~\citep{cui2024origen} is a Verilog  instruction–response corpus with 222,075 examples, used to
fine-tune LLMs for register-transfer-level (RTL) code generation.
%evaluate \xin{not for evaluate but for fine-tuning RTL code} 
 \underline{Evol-CodeAlpaca-V1}~\citep{luo2024wizardcoder} is a multi-language instruction-to-code dataset of roughly 111,000 examples (covering Python, C++, Java, TypeScript, and Shell). 

Following prior work~\citep{cui2024origen,luo2024wizardcoder}, we adopt 
\textbf{DeepSeek-Coder-7B-Instruct-v1.5}~\citep{guo2024deepseek} and \textbf{CodeLlama-7B-Instruct}~\citep{roziere2023code} as base models to validate the memorization barrier on the OriGen dataset. For Evol-CodeAlpaca-V1, we use \textbf{DeepSeek-Coder-7B-v1.5}~\citep{guo2024deepseek} and \textbf{Llama-3-8B}~\citep{grattafiori2024llama} as base models. And see \textbf{Appendix\,\ref{appendix: finetune_details}} for more details about model fine-tuning parameters for FT and IB-FT.

\vspace{2mm}
\noindent \textbf{Evaluation.} For \underline{OriGen}, we evaluate using the \textbf{VerilogEval} benchmark~\citep{liu2023verilogeval}, which comprises two subsets: \textbf{Eval-Human} (human-authored problems, \textit{e.g.}, HDLBits and manually transcribed tasks) and \textbf{Eval-Machine} (machine-generated prompts produced by LLMs). For \underline{Evol-CodeAlpaca-V1}, we use \textbf{HumanEval}~\citep{chen2021evaluating}, a standard Python benchmark of 164 programming problems with natural-language descriptions, function signatures, and unit tests for automatic correctness evaluation. For evaluation metric, we introduce Pass@$k^{(m)}$, which requires at least $m$ of $k$ samples to succeed, offering a stricter and more reliable measure of generation consistency than standard Pass@$k$. More metric details shown in \textbf{Appendix\,\ref{appendix:evaluation}}.
\subsection{Experiment Results}

\begin{table*}[!t]
\centering
\setlength{\tabcolsep}{6pt}
\renewcommand{\arraystretch}{1.2}
\caption{
Performance for DeepSeek-Coder-7B-Instruct-v1.5 and CodeLlama-7B-Instruct fine-tuned on OriGen using IB-FT versus conventional fine-tuning (FT). Test-time accuracies are reported on the Eval-Human and Eval-Machine subsets using Pass@$k$ ($k\in\{1,5,10\}$) and the stricter Pass@$10^{(m)}$ ($m\in\{2,5,10\}$), where Pass@$10^{(m)}$ counts a problem as solved only if at least $m$ of the 10 generated samples pass the unit tests. Best results are shown in \textbf{bold}.
}
\label{tab:main_verilogeval}
\begin{small}
\scalebox{0.85}{
\begin{tabular}{l|ccc|ccc|ccc|ccc}
\toprule[1pt]
\multirow{4}{*}{\textbf{Methods}}  & \multicolumn{6}{c|}{\textbf{Eval-Human (\%)}} & \multicolumn{6}{c}{\textbf{Eval-Machine (\%)}} \\
\cmidrule(lr){2-7} \cmidrule(lr){8-13}
& \multicolumn{3}{c|}{\makecell{\(\mathrm{Pass@}k\uparrow\)}} 
& \multicolumn{3}{c|}{\makecell{\(\mathrm{Pass@}10^{(m)} \uparrow\)}} 
& \multicolumn{3}{c|}{\makecell{\(\mathrm{Pass@}k\uparrow\)}} 
& \multicolumn{3}{c}{\makecell{\(\mathrm{Pass@}10^{(m)} \uparrow\)}} \\
\cmidrule(lr){2-4} \cmidrule(lr){5-7} \cmidrule(lr){8-10} \cmidrule(lr){11-13}
& $k=1$ & $k=5$ & $k=10$ & $m=2$ & $m=5$ & $m=10$ & $k=1$ & $k=5$ & $k=10$ & $m=2$ & $m=5$ & $m=10$ \\
\midrule
\multicolumn{13}{c}{\textbf{DeepSeek-Coder-7B-Instruct-v1.5}} \\
\midrule
\textbf{Base}   & 34.62 & 42.09 & 44.23 & 42.31 & 36.54 & 24.36 & 56.47 & 64.76 & 67.13 & 63.64 & 58.74 & 43.66 \\
\midrule
\textbf{+ FT} & 42.63 & 54.60 & \textbf{57.77} & 54.49 & 41.03 & 27.56 & 67.83 & 76.02 & 76.92 & 76.23 & 72.73 & 49.65 \\
\cellcolor{LightCyan!50}\textbf{+ IB-FT}  & \cellcolor{LightCyan!50}\textbf{51.99} & \cellcolor{LightCyan!50}\textbf{54.86} & \cellcolor{LightCyan!50}55.77 & \cellcolor{LightCyan!50}\textbf{54.49} & \cellcolor{LightCyan!50}\textbf{51.92} & \cellcolor{LightCyan!50}\textbf{48.07} & \cellcolor{LightCyan!50}\textbf{75.67} & \cellcolor{LightCyan!50}\textbf{78.51} & \cellcolor{LightCyan!50}\textbf{79.02} & \cellcolor{LightCyan!50}\textbf{78.32} & \cellcolor{LightCyan!50}\textbf{77.62} & \cellcolor{LightCyan!50}\textbf{69.23} \\
\midrule
\multicolumn{13}{c}{\textbf{CodeLlama-7B-Instruct}} \\
\midrule
\textbf{Base}   & 17.56 & 27.02 & 29.49 & 26.28 & 17.95 & 4.19 & 44.76 & 53.79 & 55.24 & 53.85 & 47.55 & 27.27 \\
\midrule
\textbf{+ FT} & 46.15 & 48.47 & 43.98 & 48.72 & 45.13 & 43.59 & 66.36 & 70.53 & 71.33 & 70.63 & 65.03 & 62.24 \\
\cellcolor{LightCyan!50}\textbf{+ IB-FT}  & \cellcolor{LightCyan!50}\textbf{49.68} & \cellcolor{LightCyan!50}\textbf{52.82} & \cellcolor{LightCyan!50}\textbf{46.81} & \cellcolor{LightCyan!50}\textbf{51.28} & \cellcolor{LightCyan!50}\textbf{50.64} & \cellcolor{LightCyan!50}\textbf{45.51} & \cellcolor{LightCyan!50}\textbf{70.07} & \cellcolor{LightCyan!50}\textbf{72.94} & \cellcolor{LightCyan!50}\textbf{73.47} & \cellcolor{LightCyan!50}\textbf{72.73} & \cellcolor{LightCyan!50}\textbf{70.63} & \cellcolor{LightCyan!50}\textbf{64.33} \\
\bottomrule[1pt]
\end{tabular}
}
\end{small}
\end{table*}

\noindent \textbf{Performance overview of IB-regularized fine-tuning \eqref{eq:final-loss}.} In \textbf{Table\,\ref{tab:main_verilogeval}} we present performance for LLMs ({DeepSeek-Coder-7B-Instruct-v1.5} and {CodeLlama-7B-Instruct}) fine-tuned on OriGen using FT and our IB-FT, together with each model's pre-fine-tuning performance. Test-time accuracy is reported on the Eval-Human and Eval-Machine subsets using standard $\mathrm{Pass@}k$ ($k\in\{1,5,10\}$) and the stricter $\mathrm{Pass@}10^{(m)}$ (fixed $k=10$, $m\in\{2,5,10\}$).

As Table\,\ref{tab:main_verilogeval} shows, IB-FT consistently yields the top-rank performance: it outperforms both the base model and conventional FT across nearly all Pass@$k$ settings, with especially large gains at $k=1$. For example, on DeepSeek-Coder-7B-Instruct-v1.5 IB-FT achieves 51.99\% (Eval-Human) and 75.67\% (Eval-Machine) for Pass@1, versus 42.63\% and 67.83\% for FT. Increasing $k$ (\textit{e.g.}, to $k=10$) narrows this gap, FT reaches 57.77\% Pass@10 for DeepSeek-Coder-7B-Instruct-v1.5, which is better than IB-FT (55.77\%). This suggests that some FT gains are realized only via sampling-based evaluation (Pass@$k$) rather than as reliable top-1 improvements, consistent with Fig.\,\ref{fig:motivation_limitation_FT}.
 Crucially, under the stricter $\mathrm{Pass@}10^{(m)}$ criteria ($m>1$), FT’s scores fall off sharply, indicating that its successes are often isolated single samples rather than consistently reproducible generations. By contrast, IB-FT delivers more robust and reproducible improvements (notably at Pass@$1$ and Pass@$10^{(m)}$ for $m > 1$).

To stress, under the stricter {Pass@}$10^{(m)}$ metric, IB-FT substantially outperforms conventional FT. For the strictest setting ($m=10$), which requires all 10 generations to pass, IB-FT improves over FT on DeepSeek-Coder-7B-Instruct-v1.5 by roughly 20\% on both Eval-Human (\textit{i.e.}, 48.07\% by IB-FT vs. 27.56\% by FT) and Eval-Machine (\textit{i.e.}, 69.23\% by IB-FT vs. 49.65\% by FT). This shows that IB-FT yields stronger generalization and more stable multi-sample performance under demanding criteria, producing more reliable models in practice.

In addition, \textbf{Table\, \ref{tab:main_evolve}} in \textbf{Appendix\,\ref{appendix:exp_Evol}} shows that IB-FT also consistently outperforms FT on the Evol-CodeAlpaca-V1 dataset. For example, on HumanEval, IB-FT improves Pass@1 from 56.70\% to 61.60\% for Llama-3-8B and from 51.28\% to 54.57\% for DeepSeek-Coder-7B-v1.5. Under the stricter $\mathrm{Pass@}10^{(m)}$ setting, IB-FT further demonstrates stronger robustness, maintaining higher scores than FT across all $m$ values.
Also \textbf{Fig.\, \ref{fig:remove_on_sft_exp}} in \textbf{Appendix\,\ref{appendix:exp_limitation}} shows pruning memorized examples, though a natural remedy, is highly sensitive, whereas IB-FT is more robust.

\begin{figure*}[!htb]
\centering
% ==========
\includegraphics[width=0.45\textwidth]{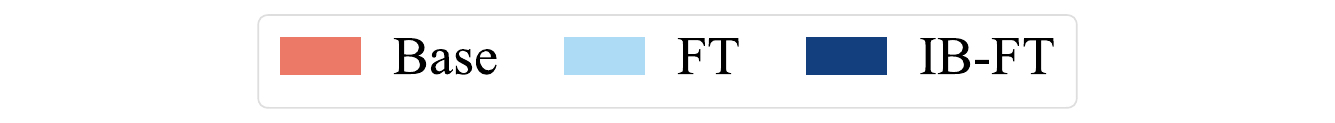} 
\vspace{-5mm}
\setlength{\tabcolsep}{0pt}
\begin{tabular}{cccc}   
  \hspace*{-1mm} \includegraphics[width=0.24\textwidth]{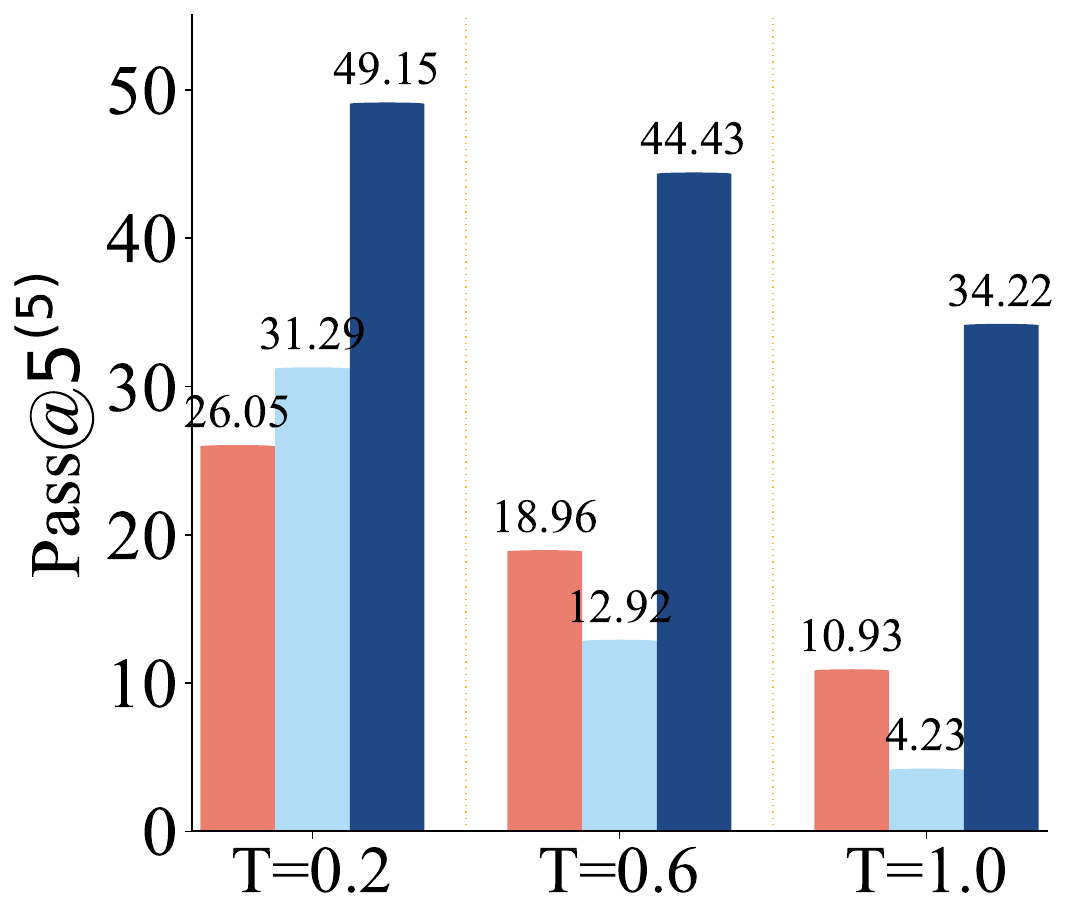}  &  
   \hspace*{1mm} 
   \includegraphics[width=0.24\textwidth]{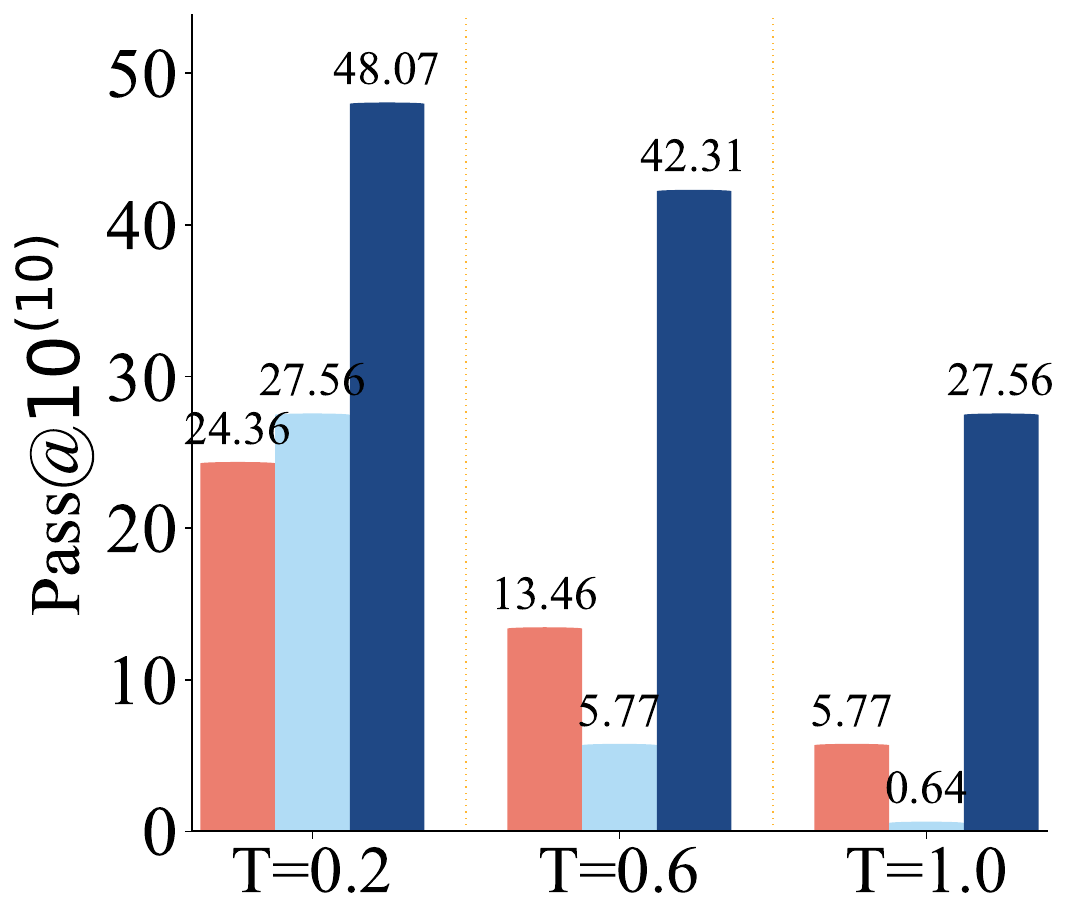} &
      \hspace*{1mm} 
   \includegraphics[width=0.24\textwidth]{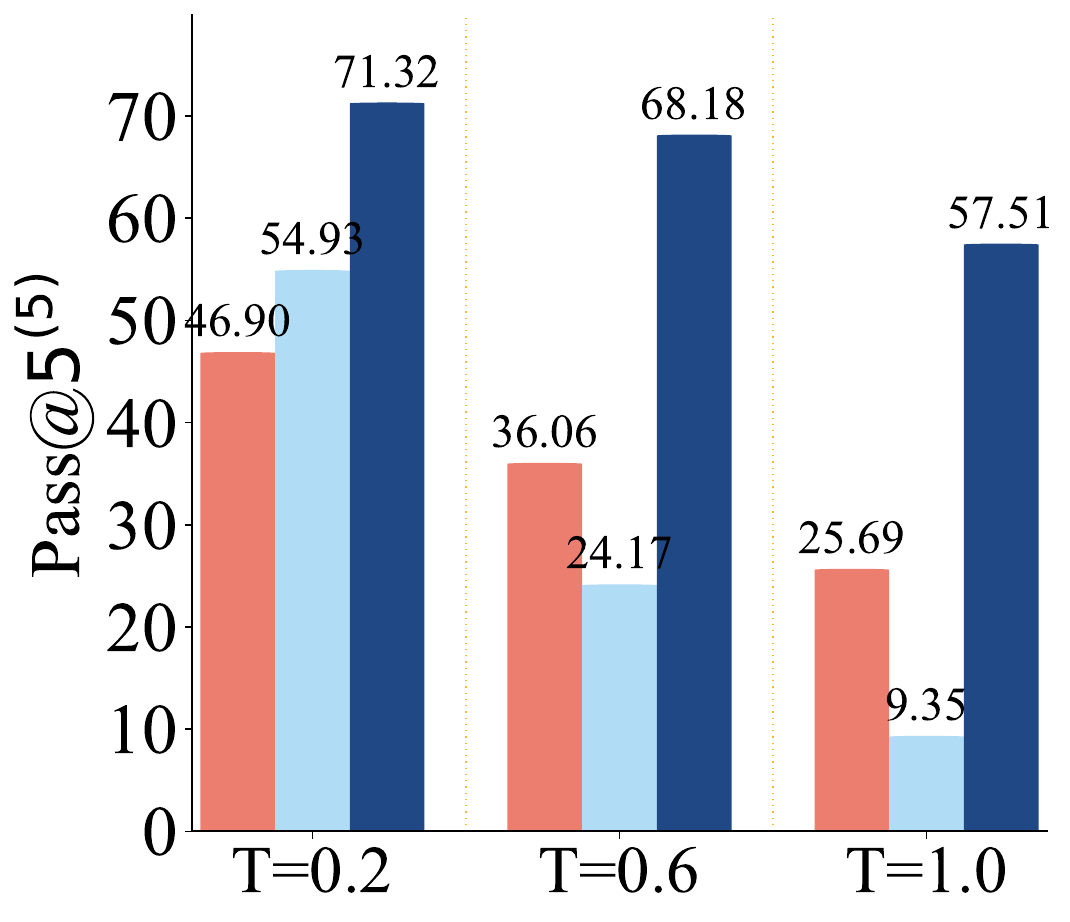}  &  
      \hspace*{1mm} 
   \includegraphics[width=0.24\textwidth]{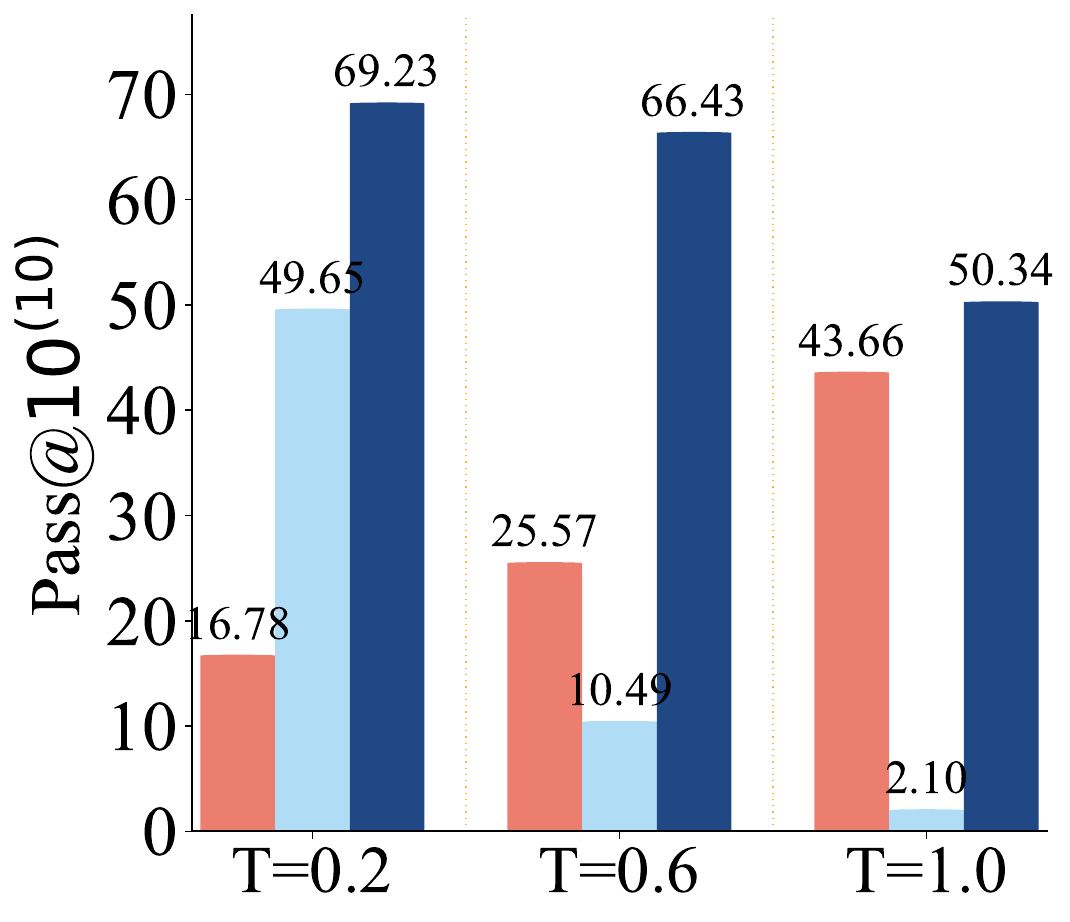} \\
   {\scriptsize (a) Eval-Human under Pass@$5^{(5)}$}  & 
       \hspace*{3mm} 
   {\scriptsize (b) Eval-Human under Pass@$10^{(10)}$}  & 
       \hspace*{3mm} 
       {\scriptsize (c) Eval-Machine under Pass@$5^{(5)}$} & 
           \hspace*{3mm} 
       {\scriptsize (d) Eval-Machine under Pass@$10^{(10)}$} \\
\end{tabular}
\vspace*{3mm}
\caption{\small{
Performance of DeepSeek-Coder-7B-Instruct-v1.5 fine-tuned with IB-FT versus conventional FT and its base model (before fine-tuning) on OriGen across decoding temperatures $T\in\{0.2,0.6,1.0\}$. Results are reported on Eval-Human and Eval-Machine using $\mathrm{Pass@}k^{(m)}$ with $m=k\in\{5,10\}$. Other experimental setups follow Table\,\ref{tab:main_verilogeval}.
}}
\label{fig:pass_temp}
\vspace*{-4mm}
\end{figure*}

\noindent \textbf{Robustness of IB-FT under different temperature ($T$) settings.} Next, we evaluate code generation after fine-tuning across decoding temperatures $T\in\{0.2,0.6,1.0\}$ (the default choice is $T=0.2$). Varying $T$ probes model robustness to sampling stochasticity and provides a more realistic assessment of fine-tuned performance and stability in practice. This protocol is motivated by the often-overlooked role of temperature in code model evaluation, since higher $T$ yields more diverse generations and can materially affect Pass@$k$-style metrics~\citep{wei2025vericoder,cui2024origen}.

\textbf{Fig.\,\ref{fig:pass_temp}} compares DeepSeek-Coder-7B-Instruct-v1.5 fine-tuned with IB-FT, the same model fine-tuned with conventional FT, and the pre-fine-tuned base model across decoding temperatures \(T\in\{0.2,0.6,1.0\}\).
Accuracy is measured by {Pass@}$k^{(m)}$ with $m=k\in\{5,10\}$, \textit{i.e.}, reflecting the all-generation success criterion if all $k$ generated samples pass the unit tests.
As we can see, IB-FT consistently sustains superior performance across all temperatures and benchmarks (Eval-Human and Eval-Machine), substantially outperforming both conventional FT and the base model even at high temperature. For example, on Eval-Human at $T=1.0$, IB-FT attains 34.22\% (Pass@$5^{(5)}$) versus 4.32\% for FT; on Eval-Machine at $T=1.0$, IB-FT reaches 50.34\% while FT collapses to 2.10\%. By contrast, FT shows poor robustness to temperature: it only exceeds the base model at the lowest temperature ($T=0.2$) and suffers large drops as $T$ increases. For instance, on Eval-Human at $T=0.6$ FT falls to 12.92\% (Pass@$5^{(5)}$) below the base model’s 18.96\%, and on Eval-Machine at $T=1.0$ the base model retains 43.66\% (Pass@$10^{(10)}$) while FT falls to 2.10\%. These results indicate that IB-FT mitigates memorization-induced fragility and yields far more stable generalization under decoding stochasticity.

\section{Conclusion}

% We uncover the memorization barrier in LLM code fine-tuning, where pretrained models already memorize many fine-tuning examples, limiting adaptation and generalization. To address this, we propose IB-regularized fine-tuning (IB-FT), which suppresses spurious memorization while preserving task-relevant signals. Experiments on OriGen and Evol-CodeAlpaca-V1 show that IB-FT consistently improves robustness and performance, especially under stricter evaluation metrics. These findings highlight IB-FT as a principled solution for overcoming memorization-driven limitations in LLM fine-tuning.

% We identify a \emph{memorization barrier} in LLM code fine-tuning: pretrained models often already memorize many downstream examples, which traps optimization and limits true adaptation and generalization. To address this, we introduce \emph{IB-regularized fine-tuning} (IB-FT), which applies an information bottleneck (IB) penalty on hidden representations to suppress spurious memorized features while preserving task-relevant signals. Across OriGen and Evol-CodeAlpaca-V1, IB-FT yields consistently stronger and more stable gains, notably improving Pass@1 and outperforming standard FT under the stricter $\mathrm{Pass@}k^{(m)}$ criterion (a problem is counted as solved only if at least $m$ of $k$ samples pass unit tests). These results position IB-FT as a practical, principled method for breaking memorization-driven limits in LLM adaptation. 

We identify a \emph{memorization barrier} in LLM code fine-tuning: pretrained models often already memorize many downstream examples, which traps optimization and limits true adaptation and generalization. To address this, we introduce \emph{IB-regularized fine-tuning} (IB-FT), which applies an information bottleneck (IB) penalty on hidden representations to suppress spurious memorized features while preserving task-relevant signals. Across OriGen and Evol-CodeAlpaca-V1, IB-FT yields consistently stronger and more stable gains, which position IB-FT as a practical, principled method for breaking memorization-driven limits in LLM adaptation.

% Future work will extend IB-FT beyond code and develop deeper theoretical understanding of IB in optimization and LLM training.

% We identify a \emph{memorization barrier} in LLM code fine-tuning: pretrained models often already memorize many downstream examples, which traps optimization and limits adaptation. To overcome this, we propose \emph{IB-regularized fine-tuning} (IB-FT), applying an information bottleneck penalty on hidden representations to suppress memorized features while preserving task signals. On OriGen and Evol-CodeAlpaca-V1, IB-FT achieves stronger and more stable gains, improving Pass@1 and outperforming standard FT under stricter $\mathrm{Pass@}k^{(m)}$.

% We identify a \emph{memorization barrier} in LLM code fine-tuning: pretrained models often already memorize downstream examples, which traps optimization and limits adaptation. To address this, we propose \emph{IB-regularized fine-tuning} (IB-FT), applying an information bottleneck penalty on hidden representations to suppress memorized features while preserving task signals. On OriGen and Evol-CodeAlpaca-V1, IB-FT delivers consistently higher accuracy than standard FT.

% We identify a \emph{memorization barrier} in LLM code fine-tuning: pretrained models memorize downstream examples, which traps optimization and limits adaptation. To address this, we propose \emph{IB-regularized fine-tuning} (IB-FT), applying an information bottleneck on representations to suppress memorized features while preserving task signals. On OriGen and Evol-CodeAlpaca-V1, IB-FT delivers consistently higher performance than FT.

\section*{Acknowledgment}
%\vspace*{-3mm}
We thank Intel for providing computing resources. The work of C. Wang and S. Liu was partially supported by the National Science Foundation (NSF) CAREER Award IIS-2338068.

% \input{section/Limitation}

% \nocite{*}
\bibliography{refs/RA,refs/MU,refs/MU_SLiu,refs/xin_ref}

\bibliographystyle{iclr2025_conference}

% \clearpage
% \newpage

% \section{Appendix}
% \label{sec:appendix}

\clearpage
\onecolumn
\section*{\Large{Appendix}}
\setcounter{section}{0}
\setcounter{figure}{0}
\setcounter{table}{0}
\makeatletter 
\renewcommand{\thesection}{\Alph{section}}
\renewcommand{\theHsection}{\Alph{section}}
\renewcommand{\thefigure}{A\arabic{figure}}
\renewcommand{\theHfigure}{A\arabic{figure}}
\renewcommand{\thetable}{A\arabic{table}}
\renewcommand{\theHtable}{A\arabic{table}}
\makeatother

\renewcommand{\thetable}{A\arabic{table}}
\setcounter{mylemma}{0}
\renewcommand{\themylemma}{A\arabic{mylemma}}
\setcounter{equation}{0}
\renewcommand{\theequation}{A\arabic{equation}}

\section{Experiment Setup and Implementation Details}
\label{appendix: setup}

\subsection{Fine-tuning Details}
\label{appendix: finetune_details}

% We employ the LoRA~(Low-Rank Adaptation)\citep{hu2022lora} method to fine-tune models across different configurations.

\paragraph{Base model.} For reproducibility, we explicitly list the pretrained code foundation models used for fine-tuning on each dataset with link. On the OriGen dataset, we fine-tune DeepSeek-Coder-7B-Instruct-v1.5\footnote{\url{https://huggingface.co/deepseek-ai/deepseek-coder-7b-instruct-v1.5}} and CodeLlama-7B-Instruct\footnote{\url{https://huggingface.co/codellama/CodeLlama-7b-Instruct-hf}}.  On the Evol-CodeAlpaca-V1 dataset, we fine-tune DeepSeek-Coder-7B-v1.5\footnote{\url{https://huggingface.co/deepseek-ai/deepseek-coder-7b-base-v1.5}} and Meta-Llama-3-8B\footnote{\url{https://huggingface.co/meta-llama/Meta-Llama-3-8B}}.

\paragraph{Fine-tuning setup for FT.}
For the baseline FT, we employ the LoRA~(Low-Rank Adaptation) method~\citep{hu2022lora}. Across all datasets and models, we adopt the AdamW optimizer with $\beta_1=0.9$ and $\beta_2=0.999$, cosine learning rate decay, and a warm-up ratio of 0.03. Training is performed for 3 epochs with a batch size of 4, using mixed \texttt{float16} precision (while the base models were pretrained in \texttt{bfloat16}). The LoRA configuration is fixed to $r=32$ and a dropout rate of 0.05, while $\alpha$ varies by dataset. On the OriGen dataset, training is conducted on two NVIDIA A100 GPUs with 80\,GB memory, using an initial learning rate of $5 \times 10^{-5}$ and LoRA $\alpha=32$.  
On the Evol-CodeAlpaca-V1 dataset, training is conducted on four NVIDIA H100 GPUs, with an initial learning rate of $1 \times 10^{-4}$ and LoRA $\alpha=64$.

\paragraph{Fine-tuning setup for IB-FT.}
Our proposed IB-FT method is also built on the LoRA~(Low-Rank Adaptation) framework. Across all datasets, we adopt the AdamW optimizer with $\beta_1=0.9$ and $\beta_2=0.999$, cosine learning rate decay, and a warm-up ratio of 0.03. Training is performed for 3 epochs with a batch size of 4, using mixed \texttt{float16} precision (while the base models were pretrained in \texttt{bfloat16}). For LoRA configuration, we set the rank $r=32$ and the dropout rate to 0.05, while $\alpha$ varies depending on the dataset. On the OriGen dataset, IB-FT is trained on two NVIDIA A100 GPUs with 80\,GB memory, using an initial learning rate of $5 \times 10^{-5}$ and LoRA $\alpha=32$.  
On the Evol-CodeAlpaca-V1 dataset, IB-FT is trained on NVIDIA H100 GPUs, with an initial learning rate of $1 \times 10^{-4}$ and LoRA $\alpha=64$.  

In addition to the above configurations, IB-FT introduces two extra hyperparameters in Eq.\,\eqref{eq:final-loss}: the IB regularization strengths $\alpha$ and $\beta$, which balance the trade-off between compression and prediction. We search over $\alpha \in (0.01, 1)$ and $\beta \in (0.001, 0.1)$. On the OriGen dataset, we set $\alpha=0.1$ and $\beta=0.02$ for both models, while on the Evol-CodeAlpaca-V1 dataset we use $\alpha=0.1$ and $\beta=0.01$.

\subsection{Evaluation Metrics} 
\label{appendix:evaluation}

In addition to the standard $\mathrm{Pass@}k$ metric, we adopt a stricter variant $\mathrm{Pass@}k^{(m)}$ (as introduced in Fig.\,\ref{fig:motivation_limitation_FT}). Under $\mathrm{Pass@}k^{(m)}$, a problem is counted as solved only if at least $m$ out of the $k$ generated samples pass the unit tests (so $\mathrm{Pass@}k^{(1)}$ equals the conventional $\mathrm{Pass@}k$). By requiring $m>1$, $\mathrm{Pass@}k^{(m)}$ penalizes lucky single-sample successes and provides a more reliable measure of a model's consistent generation quality and robustness. To evaluate this metric, we need to generate a pool of $n$ candidate outputs for each problem (with $n \geq k$). Among these $n$ outputs, the parameters $k$ and $m$ are then defined relative to this pool: $k$ is the subset size under evaluation, and $m$ is the minimum number of successful outputs required within that subset. In our experiments, we fix $n=10$.
\section{Additional Experiment Results}
\label{appendix: results}

\subsection{Performance of Fine-tuning  on Evol-CodeAlpaca-V1} 
\label{appendix:exp_Evol}

We also evaluate {Llama-3-8B} and {DeepSeek-Coder-7B-v1.5} on Evol-CodeAlpaca-V1, comparing conventional fine-tuning (FT), IB-FT, and the base model before fine-tuning. Evaluation is carried out on the HumanEval, using both the standard $\mathrm{Pass@}k$ metric ($k\in{1,5,10}$) and the more demanding $\mathrm{Pass@}10^{(m)}$ criterion (with $k=10$, $m\in{2,5,10}$). 

As \textbf{Table\,\ref{tab:main_evolve}} shows, IB-FT consistently provides the strongest results across both models, with particularly notable improvements at $k=1$. For instance, on Llama-3-8B, IB-FT achieves 61.60\% for Pass@1 compared to 56.70\% for FT, while on DeepSeek-Coder-7B-v1.5 the gains are 54.57\% versus 51.28\%. Moreover, under the strict $\mathrm{Pass@}10^{(m)}$ setting, IB-FT demonstrates substantially higher robustness, reaching 37.20\% at $m=10$ on Llama-3-8B, which still provides a clear improvement over FT’s 30.49\%. These results highlight that IB-FT not only enhances top-1 accuracy but also yields more consistent and reproducible improvements under demanding multi-sample criteria, underscoring its advantage over conventional fine-tuning.

\begin{table*}[htb]
\centering
\setlength{\tabcolsep}{6pt}
\renewcommand{\arraystretch}{1.2}
\caption{Performance overview for Llama-3-8B and DeepSeek-Coder-7B-v1.5 fine-tuned on Evol-CodeAlpaca-V1 using IB-FT versus conventional fine-tuning (FT). Other metrics and table format follow those in Table\,\ref{tab:main_verilogeval}.}
\label{tab:main_evolve}
\begin{small}
\scalebox{0.85}{
\begin{tabular}{l|ccc|ccc}
\toprule[1pt]
\multirow{4}{*}{\textbf{Methods}} & \multicolumn{6}{c}{\textbf{HumanEval (\%)}} \\
\cmidrule(lr){2-7}
& \multicolumn{3}{c|}{\makecell{\(\mathrm{Pass@}k \uparrow\)}} 
& \multicolumn{3}{c}{\makecell{\(\mathrm{Pass@}10^{(m)} \uparrow\)}} \\
\cmidrule(lr){2-4} \cmidrule(lr){5-7}
& $k=1$ & $k=5$ & $k=10$ & $m=2$ & $m=5$ & $m=10$ \\
\midrule
\multicolumn{7}{c}{\textbf{Llama-3-8B}} \\
\midrule
\textbf{Base}   & 29.09 & 41.22 & 46.95 & 37.80 & 28.66 & 17.07 \\
\textbf{+ FT}   & 56.70 & 74.16 & 79.27 & 73.17 & 62.80 & 30.49 \\
\cellcolor{LightCyan!50}\textbf{+ IB-FT} 
                & \cellcolor{LightCyan!50}\textbf{61.60} & \cellcolor{LightCyan!50}\textbf{75.44} & \cellcolor{LightCyan!50}\textbf{79.88} 
                & \cellcolor{LightCyan!50}\textbf{73.78} & \cellcolor{LightCyan!50}\textbf{65.24} & \cellcolor{LightCyan!50}\textbf{37.20} \\
\midrule
\multicolumn{7}{c}{\textbf{DeepSeek-Coder-7B-v1.5}} \\
\midrule
\textbf{Base}   & 44.33 & 56.89 & 62.20 & 54.88 & 45.12 & 30.49 \\
\textbf{+ FT}   & 51.28 & 64.10 & 64.10 & 63.46 & 50.00 & 32.05 \\
\cellcolor{LightCyan!50}\textbf{+ IB-FT} 
                & \cellcolor{LightCyan!50}\textbf{54.57} & \cellcolor{LightCyan!50}\textbf{67.30} & \cellcolor{LightCyan!50}\textbf{70.73} 
                & \cellcolor{LightCyan!50}\textbf{67.07} & \cellcolor{LightCyan!50}\textbf{54.88} & \cellcolor{LightCyan!50}\textbf{39.63} \\
\bottomrule[1pt]
\end{tabular}
}
\end{small}
\end{table*}

\subsection{Limitation of Pruning Memorized Examples, then Fine-tuning}
\label{appendix:exp_limitation}

\begin{figure}[!htb] 
\vspace*{0mm}
\centering  
\begin{tabular}{cc}
%\hspace*{-6mm}
\includegraphics[width=0.3\textwidth]{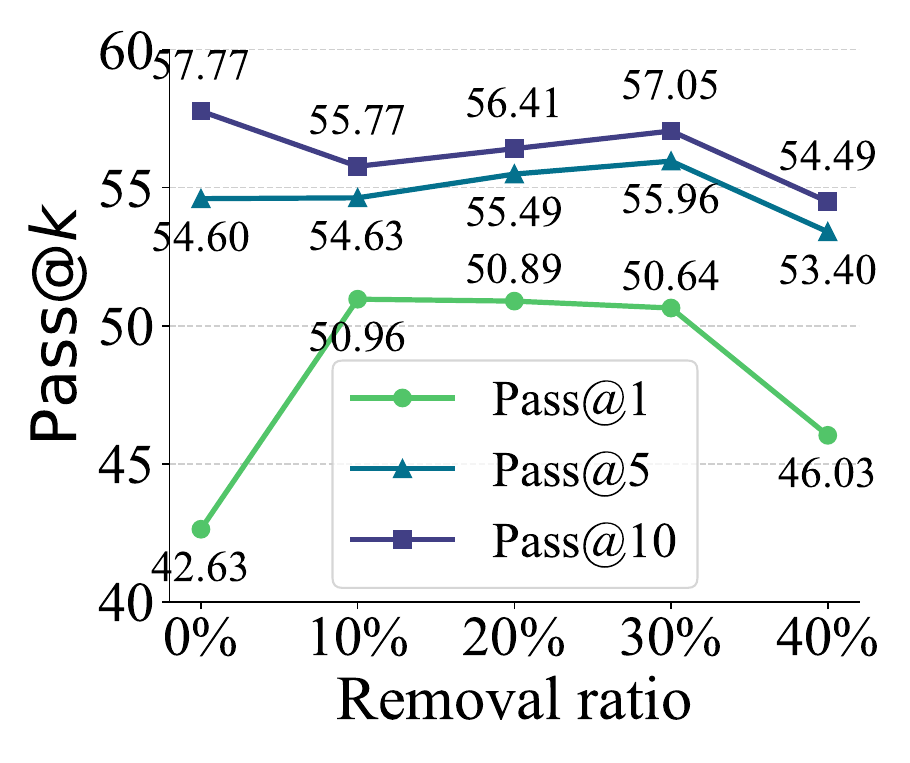} &
%\hspace*{-4mm}
\includegraphics[width=0.3\textwidth]{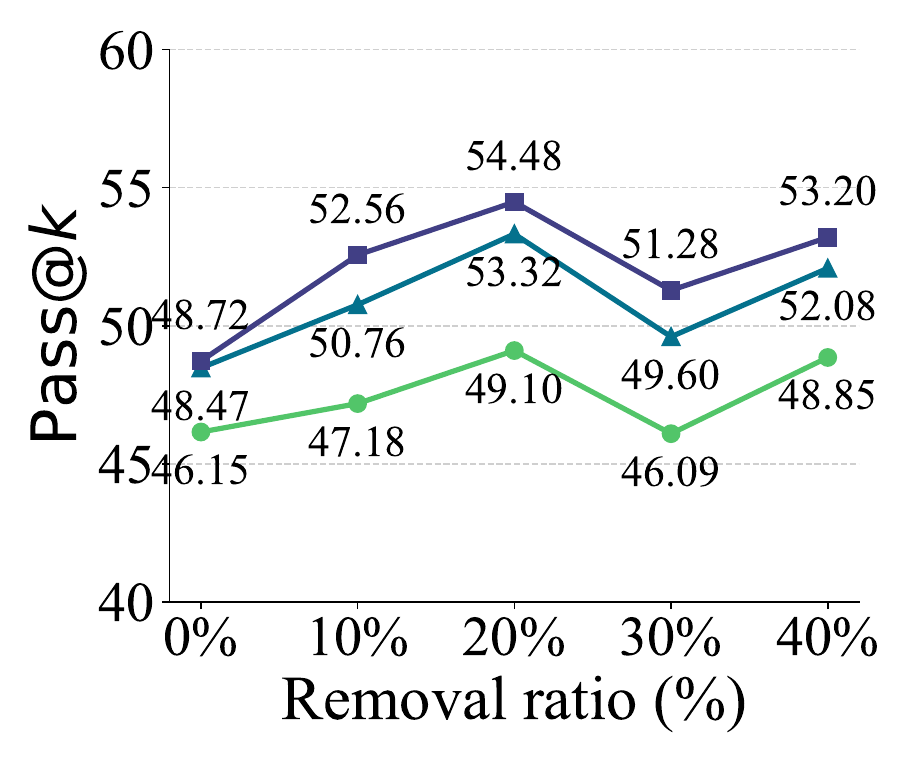} \vspace*{-2mm}\\
  {\small (a) DeepSeek-Coder-7B-Instruct-v1.5}  &
  {\small (b) CodeLlama-7B-Instruct} 
\end{tabular}
\vspace*{-0mm}
\caption{\small{Performance of fine-tuning with different data-removal ratios on OriGen, evaluated on the VerilogEval (Eval-Human) subset. Fine-tuning is applied to the base model (a) {DeepSeek-Coder-7B-Instruct-v1.5} and (b) {CodeLlama-7B-Instruct}, and the performance is assessed using \text{Pass@}$k$ ($k\in\{1,5,10\}$).
}
}
\label{fig:remove_on_sft_exp}
\vspace*{-4mm}
\end{figure}

Recall from Sec.\,\ref{sec:Method1} that removing highly-memorized examples (as characterized by \texttt{Min-K\% Prob} scores) is a natural countermeasure to memorization. In \textbf{Fig.\,\ref{fig:remove_on_sft_exp}} we further examine this strategy on two base models (DeepSeek-Coder-7B-Instruct-v1.5 and CodeLlama-7B-Instruct) by removing $10\%$–$40\%$ of the most-memorized examples and reporting Pass@$k$ for $k\in\{1,5,10\}$. The results reveal two sensitivity dimensions. First, the optimal removal ratio differs across models (\textit{e.g.}, DeepSeek-Coder-7B-Instruct-v1.5 peaks at $10\%$ for Pass@1 while CodeLlama-7B-Instruct peaks near $20\%$). Second, the optimal ratio varies with the metric even for the same model (\textit{e.g.}, DeepSeek-Coder-7B-Instruct-v1.5’s Pass@1 peaks at $10\%$ whereas Pass@5 peaks at $30\%$). As shown, there is no universal pruning rate: removal is model- and metric-dependent. By contrast, our proposal avoids fragile data pruning and can be compatible with memorized examples during fine-tuning.

\clearpage
\newpage

\end{document}